\documentclass[twoside,11pt]{article} 
\usepackage{jmlr2e_cm} 

\usepackage{bm}

\usepackage{mathrsfs }

\newcommand{\x }{\bm{x}}

\usepackage{amsmath,amssymb,amsxtra,comment,graphicx,psfrag}
\usepackage{bm,mathrsfs}
\usepackage{mathtools}
\usepackage{xspace}
\usepackage{stmaryrd}
\usepackage{color}

\usepackage{enumerate,comment,graphicx,psfrag}

\usepackage{hhline}

\usepackage{bookmark}
\usepackage{nameref}

\usepackage{verbatim}
\usepackage{tabls}

\usepackage{fancyhdr}
\usepackage{epsfig}
\usepackage{epsfig,subfigure,epstopdf}

\include{rgb}

\def\nst2{\| _*} 
 
\def\a12{A_h ^{1/2} } 
\def\d{{\mathrm d}}
\def\tr|{|\! |\! |}

\def\R {{\mathbb R}}
\def\P{{\mathbb P}}

\def\E{{\mathcal{E}}}

\def \a{\alpha }

\def\T_h{{{\mathcal T}_h}}

\def\<{{\langle }}
\def\>{{\rangle }}

\DeclarePairedDelimiter{\norm}{\lVert}{\rVert}

\DeclareSymbolFont{matha}{OML}{txmi}{m}{it}
\DeclareMathSymbol{\varv}{\mathbf}{matha}{118}

\newtheorem{prop}{Proposition}[section]

\newtheorem{Theorem}{Theorem}[section]

\newtheorem{Lemma}{Lemma}[section]
\newtheorem{Proposition}{Proposition}[section]

\newcommand\restr[2]{{
  \left.\kern-\nulldelimiterspace 
  #1 
  \vphantom{\big|} 
  \right|_{#2} 
  }}

\NewDocumentCommand{\dgal}{sO{}m}{%
  \IfBooleanTF{#1}
    {\dgalext{#3}}
    {\dgalx[#2]{#3}}%
}

\NewDocumentCommand{\dgalext}{m}{%
  \sbox0{%
    \mathsurround=0pt 
    $\left\{\vphantom{#1}\right.\kern-\nulldelimiterspace$%
  }%
  \sbox2{\{}%
  \ifdim\ht0=\ht2
    \{\kern-.625\wd2 \{#1\}\kern-.625\wd2 \}%
  \else
    \left\{\kern-.7\wd0\left\{#1\right\}\kern-.7\wd0\right\}%
  \fi
}

\NewDocumentCommand{\dgalx}{om}{%
  \sbox0{\mathsurround=0pt$#1\{$}%
  \sbox2{\{}%
  \ifdim\ht0=\ht2
    \{\kern-.625\wd2 \{#2\}\kern-.625\wd2 \}%
  \else
    \mathopen{#1\{\kern-.7\wd0 #1\{}
    #2
    \mathclose{#1\}\kern-.7\wd0 #1\}}
  \fi
}

\renewcommand{\R}{{\mathbb{R}}}

\renewcommand{\T}{\mathbb{T}^d}

\newcommand{\PP}{\mathbb{P}}
\newcommand{\EE}{\mathbb{E}}

\newcommand{\be}{\begin{equation}}
\newcommand{\ee}{\end{equation}}

\numberwithin{equation}{section}
\numberwithin{table}{section}
\numberwithin{figure}{section}

\ShortHeadings{}{}
\firstpageno{1}

\begin{document}


\title
{A new approach to generalisation error of  machine learning algorithms: Estimates and convergence} 

\author{\name  Michail Loulakis \email {loulakis@mail.ntua.gr} \\
\addr 
National Technical University of Athens  
\\IACM-FORTH 
\AND
\name  Charalambos G.\ Makridakis  \email{c.g.makridakis@iacm.forth.gr} \\
\addr IACM-FORTH  \\
\addr   University of
Crete\\    
University of Sussex
} 

\editor{}
\maketitle
\abstract{In this work we consider a model problem of deep neural learning, namely the learning of a given function  when it is assumed that we have access to its point values on a finite set of points.  The \emph{deep neural network interpolant} is the the resulting approximation of $f$, which is obtained by a typical machine learning algorithm involving a given DNN architecture and an optimisation step, which is assumed  to be solved exactly. These are among the simplest regression algorithms based on neural networks.  In this work we introduce a new approach to the estimation of the (generalisation) error and to convergence. Our results include (i) estimates of the error without any structural assumption on the neural networks and under mild regularity assumptions on the learning function $f$  (ii)  convergence of the approximations to the target function $f$ by only requiring that the neural network spaces have appropriate approximation capability.}




\section{Deep neural network interpolation}\label{Se:1}

\subsection{The model problem}\label{SSe:1.1}
In this work we consider a model problem of deep neural learning, namely the learning of a given function $f : D \to  \R ,$ $D\subset \R^d,$ when we only have access to its values on a finite set of points (\emph{interpolation points}).  The \emph{deep neural network  interpolant} is the resulting approximation of $f$, which is obtained by a typical machine learning algorithm involving a given deep neural network  (DNN) architecture and an optimisation step.{ For convenience, we assume that the minimisation problem can be solved exactly and we discuss the case of local minimisers in Section \ref{subsec:local_min}}. These are among the simplest regression algorithms based on neural networks.  Our aim is to show convergence of such algorithms as the number of parameters of the neural network and the number of the interpolation points both tend to infinity (in an appropriate sense).  
This problem can be  associated to the  limiting   functional
\begin{equation*}
\mathcal{E} (u)=
\int_{D}    |  u - f|^p    \d \mu(x)  \, ,
\end{equation*}
and the discrete loss is a Monte-Carlo approximation of this integral.  
Our approach can be extended to other functionals  including appropriate regularisations. 
In this work we introduce a new approach to the estimation of the (generalisation) error and to convergence. Our results include   estimates of the error without any structural assumption on the neural networks and under mild regularity assumptions on the learning function $f.$ Our analysis includes 	convergence results   of the approximations which provide qualitative information for the behaviour and the robustness of the algorithms. We show convergence to the target function $f$ by only requiring that the neural network spaces have appropriate approximation capability. 
%
%
These  results rely, as it is natural, on the ability of the DNN space to  approximate functions with sufficient regularity and to the existence of rich enough training data points.  Furthermore, we demonstrate that these two main  discretisation sources  influence the behaviour of the algorithm in a rather  {decoupled} and robust way.  
%
%

\subsection{Discrete spaces generated by DNN}\label{Se:1NN}
We need to define first the finite dimensional spaces 
generated by deep neural networks. 
%
We adopt  standard notation, see e.g.,  \cite{Mishra_Schwab_QMC_DNN:2021, Mishra_dispersive:2021}, by considering functions 
$u_\theta$ defined through neural networks.
A deep neural network maps every point $x\in D\subset \R ^d$ to a number $u_\theta (x) \in \R$.  {To fix ideas, we discuss the case of a multilayer perceptron, but the reader should bear in mind that our results depend on approximation properties of the networks, rather than their precise architecture, hence they have a much larger scope.}
\begin{equation}\label{C_L}
	u_\theta(x)= C_L  \circ \sigma  \circ C_{L-1} \cdots \circ\sigma \circ C_{1} (x) \quad \forall x\in D.
\end{equation}
%
 The process
 \begin{equation}
 \mathcal{C}_L:= C_L  \circ \sigma  \circ C_{L-1} \cdots \circ\sigma \circ C_{1} 
\end{equation}
is in principle a map $\mathcal{C}_L : \R ^m \to \R ^{m'} $; in our particular application, $m =d$ and $m'=1.$ 
The map $\mathcal{C}_L $ corresponds to a neural network with $L$ layers and activation function $\sigma.$ Notice that to 
define $u_\theta(x) $ for all $x\in D$ we use the same $\mathcal{C}_L ,$ thus $u_\theta(\cdot ) =\mathcal{C}_L  (\cdot ) .$
Any such map $\mathcal{C}_L$ is characterized by the intermediate (hidden) layers $C_k$, which are affine maps of the form 
\begin{equation}\label{C_k}
	C_k y = W_k y +b_k, \qquad \text{where }  W_k \in \R ^ {d_{k+1}\times d_k}, b_k \in \R ^ {d_{k+1}}.
\end{equation} 
Here the dimensions $d_k$ may vary with each layer $k$ and $\sigma (y)$ denotes the vector with the same number of components as $y$,
where $\sigma (y)_i= \sigma(y_i)\, .$ 
The index $\theta$ represents collectively all the parameters of the network $\mathcal{C}_L,$ namely $W_k , b_k, $ $k=1, \dots, L .$ 
The set of all networks $\mathcal{C}_L$ with a given structure (fixed $L, d_k,  k=1, \dotsc, L\,$) of the form \eqref{C_L}, \eqref{C_k}
is called $\mathcal{N}.$ The total dimension (total number of degrees of freedom) of  $\mathcal {N} ,$ is $\dim{\mathcal {N}}= \sum _{k=1} ^L d_{k+1} (d_k +1) \, .$ 
We now define the space of functions 
\begin{equation}
	V _{\mathcal{N}}= \{ u_\theta : D \to \R ,  \ \text{where }  u_\theta (x) = \mathcal{C}_L (x), \ \text{for some } \mathcal{C}_L\in  \mathcal{N}\, \} \, .
\end{equation}
It is important to observe that $V _{\mathcal{N}}$ is not a linear space. In fact, if 
$u_\theta, u_{\tilde\theta}\in V _{\mathcal{N}}$, then this means that there are $\mathcal{C}_L$ and ${\tilde {\mathcal{C}}}_L$ such that 
$u_\theta (x) =\mathcal{C}_L(x) $ and $u_{\tilde\theta} (x) = {\tilde {\mathcal{C}}}_L(x).$ This, in general, does not imply that there 
is a ${\hat   {\mathcal{C}}}_L$ such that 
$u_\theta (x) + u_{\tilde\theta} (x) = {\hat {\mathcal{C}}}_L(x).$ On the other hand, there is a one-to-one correspondence 
\begin{equation}
	\theta \mapsto  u_\theta \in V _{\mathcal{N}}\, . 
\end{equation}
We denote by 
\begin{equation}
	\Theta = \{ \theta \, : u_\theta \in V _{\mathcal{N}}\}.
\end{equation}  
Clearly, $\Theta $ is a linear subspace of $ \R ^{\dim{\mathcal {N}}}.$

It will be useful to consider  the scheme which assumes the knowledge of $f$ and thus only has theoretical scope:

\begin{definition}\label{abstract_mm_nn} Assume that the problem 
\begin{equation}\label{mm_nn:abstract}
	\min  _ {v \in  V _{\mathcal{N}} } \E (v)   
\end{equation}
where
\begin{equation*}
\mathcal{E} (u)=
\int_{D}    |  u - f|^p    \d x  , \qquad 1\leq p < \infty\, ,  
\end{equation*}
has a solution $v^\star \in V _{\mathcal{N}} .$ We call  $ v^\star \,  $ a $\mathcal{N}-$DNN minimiser of $\E \, .$
\end{definition}
This  problem is equivalently formulated as a minimisation problem in $\R ^{\dim{\mathcal {N}}}:$
\begin{equation}\label{mm_nn:abstract_theta}
	\min _ { \theta \in  \Theta } \E(u_\theta).
\end{equation}
%


\subsection{Training and MC deep neural network interpolant} The training of the algorithm assumes knowledge of $f$ only at a finite set of points 
 $x_i,$ $i=1,\dots, N $,  whose  location is either not controlled by us or chosen randomly.  We define the discrete functional 
\begin{equation}
\label{E_h}
\mathcal{E}_{N}( g )  =  \frac 1N  \sum _{i}   \Big  | g(x_i) - f(x_i) \Big  | ^p  \, ,  \qquad 1\leq p < \infty\,   . 
\end{equation}

\begin{definition} \label{D-dnnint}{\sc [deep neural network interpolation]}  
Assume that the problem 
\begin{equation}\label{ieE-minimize_NNQ}
	\min  _ {v \in  V _{\mathcal{N}} } \E _ {N}(v)
\end{equation}
has a solution $u_{\x}  \in V _{\mathcal{N}} .$ We call  $ u_{\x}  \,  $ a $V_{\mathcal{N}}-$\emph{deep neural network interpolant of $f$ at $ \x = (x_1, \dots, x_n )  \, .$}
\end{definition}

As mentioned,   the space $V_{\mathcal{N}}$ is non-linear and the above problem is solved   and    equivalently formulated as a minimisation problem in $\R ^{\dim{\mathcal {N}}}$ with respect of the parameters $\theta :$
\begin{equation}\label{mm_nn:abstract_theta2}
	\min _ { \theta \in  \Theta } \E_ {N} (u_\theta).
\end{equation}
Although \eqref{E_h} is convex in $g,$  the convexity is lost with respect to $\theta$ and thus \eqref{mm_nn:abstract_theta2} 
might have more than one minimisers. We investigate the limit behaviour 
of the minimisers as $\ell \to \infty $, meaning that both $| \mathcal {N}   | $ and $ N $ tend to $\infty\, , $ in a sense which will be made precise below. 

\section{State of the art and results} The problem of studying the behaviour of approximations $u_{\x}  \in V _{\mathcal{N}} $ is central in scientific machine learning. 
Although we neglect the effect of error on  the approximation of the minimisation problem (through, for example, stochastic gradient methods), the behaviour of  $u_{\x} $ as $N\to \infty$ and/or the 
approximation capacity of ${\mathcal{N}} $  becomes richer as the number of parameters grow will be instrumental at two fronts: (i) further understanding how the simple machine learning algorithms work and (ii) contribute to the assessment of their reliability by providing tools to control the error beyond the training sets, i.e., controlling the generalisation error.  \\

The problem in \eqref{ieE-minimize_NNQ}
 is typically modelled in a probabilistic sense. In fact, one considers
    a sequence $X_1,X_2,\ldots$ of i.i.d.\ $D$-valued random variables, which 
 model the selection of the points $x_1, x_2, \ldots .$ Then we 
 minimise the energy 
\begin{equation*}\label{prob_E}
\mathcal{E}_{N, \omega}( g )  =  \frac 1N  \sum _{i}   \Big  | g(X_i(\omega)\, ) - f(X_i (\omega) \, ) \Big  | ^p  \, , 
\end{equation*}
over the DNN space and denote a minimiser by $ u_{\bm{X} (\omega)   } ,$ see the next section for precise notation. 
We are therefore  interested to study the behaviour of the generalisation error:
$$ \|  f- u_{\bm{X}     }\| _p\, .$$
It is natural to expect that the two sources of error within the algorithm 
will influence the final control of the error: The error due to the minimisation over a DNN discrete space instead of the  $L^p, $
and the error due to the approximation of the continuous functional which is associated to $f$ by the sample avareges: 
\begin{equation*}\label{prob_E_error}
 {E}_{MC} (u_{\bm{X} (\omega)    }) =  \int_{D}    |  u_{\bm{X}  (\omega)   } - f|^p    \d x -  \frac 1N  \sum _{i}   \Big  | u_{\bm{X} (\omega)   } (X_i(\omega)\, ) - f(X_i (\omega) \, ) \Big  | ^p  \, . 
\end{equation*}
A key technical issue here is that since $u_{\bm{X} (\omega)   }$ is a solution of a minimisation problem involving $( X_i ) ,$ it obviously depends on  $( X_i ) ,$ and thus the law of large numbers and other standard  tools estimating ${E}_{MC}  ( g) $ for fixed $g$ are not applicable, see \cite{Kallenberg_book}. 
A way to overcome this issue is by estimating 
\begin{equation*}\label{prob_E_error_2}
\big | {E}_{MC}(u_{\bm{X}     }) \big | \leq \sup _ {g \in  V _{\mathcal{N}} }   \, \big | {E}_{MC}(g ) \big |\, ,   
\end{equation*}
see \cite{Bartlett_Montanari_Rakhlin_actaN:2021}. 
Then, (cf. \cite{Bartlett_Montanari_Rakhlin_actaN:2021}) the Rademacher complexity of a function class $\mathcal G$ (called $R_ {N, \mathcal G } $) is known to control $\sup _ {\mathcal G }   \, \big | {E}_{MC}(g ) \big |\, .$ 
Several works are devoted to the study of rates of $R_ {N, \mathcal G } $ 
or the  empirical control of the generalisation error through $R_ {N, \mathcal G } $, in   given sets of neural networks which comply with certain structural assumptions needed to control $R_ {N, \mathcal G } $ on   sets  $\mathcal G\, ,$
see e.g.,  \cite{Bartlett_Montanari_Rakhlin_actaN:2021} and its references, and also \cite{Bartlett_VCdimension:2019, Oneto_local_Rademacher:2018,Yousefi_local_Rademacher:2018,
	Bartlett_Benignoverfitting:2020,
	WE_Rademacher:2020,
	Dimakis_ExactlyComputingtheLocalLipschitz:2020, 
	Montanari_neural_kernel:2021}.
	Although different,  the method in 	\cite{Jentzen_full_err:2022} relies as well on structural assumptions on  the neural network spaces considered.
	As it is natural, the approach described requires neural network spaces to 
	be chosen with care so that certain properties are valid throughout 
	the function class $V _{\mathcal{N}}  \, .$ One of these is, typically, a 
	common bound of the Lipschitz constant within the elements of the class. \\
	
	In this work we suggest an alternative point of view on 
	 the generalisation error and the convergence analysis of these algorithms.  Our results aim (i) to provide 
	 generalisation error control, under as weak assumptions as possible, and
	 (ii) to provide a detailed convergence analysis of the algorithms under even weaker 
	 hypotheses. These  contributions are described below.
	 
	 \subsubsection* {A new approach to generalisation error} To derive the error bounds we utilise two key ingredients: the representation of the discrete loss through empirical measures, and the estimate of the  terms involved through Wasserstein distances. Our estimates provide both, control of the generalisation error for each given instance of the application of the algorithm, i.e., for each choice of sample points $(X_i),$ and control of the expected value over all possible samples. 
	  Both results, detailed in Theorem \ref{Thm:estimate}, assume that the target function $f$ is Lipschitz.  
	 The estimates depend on  the Lipschitz constant of the minimiser $u_{\bm{X} (\omega)   }$, and \emph{not on a uniform constant of members of the entire set} $V _{\mathcal{N}}  \, .$  We do not make any structural assumption on  $V _{\mathcal{N}}  \, .$ The bounds hinge on the optimal transport framework and on the control of the Wasserstein distance of the measures involved. They open the way for new empirical estimation of generalisation error as well as for new asymptotic estimates. In fact, an interesting feature of the bounds is their  relationship  to  the optimal matching problem in probability, see for example, \cite{AjtaiKT:1984, Talagrand:1994,  Parisi:matching:2014, FournierGuillin:2015, AmbrosioST:2019}.  Thus,  known results can be applied to yield asymptotic rates, see Theorem \ref{Thm:estimate}. Furthermore,  more refined and  problem adapted
	 estimates are possible, and interesting questions for future research arise.

	 \subsubsection*{Convergence under minimal assumptions} We undertake the task of studying the convergence behaviour of $u_{\bm{X} (\omega)   }$ to $f$ under the sole assumption that the neural network spaces can approximate functions in $L^p,$
	 see  \eqref {w_ell_7} for the precise assumption. Furthermore, we provide useful qualitative information on the behaviour of the learning algorithm by studying the limit of training samples $N\to \infty$ within a fixed DNN architecture. 
	 In fact, we provide results demonstrating (i) the minimisers converge to a limiting function which satisfies optimal approximation bounds as the limit of training samples $N\to \infty,$ while we held  fixed the DNN architecture (ii) local minimisers 
	 also have a stable behaviour, provided that their losses are bounded and (iii)
	   a.s.\ with respect to the sampling parameter, 
the sequence 
$(u_{\bm {X}_{N(\ell)} (\omega)} )_\ell$ converges     to $f,$ see Theorem  
	\ref{calNtoinfty_thm}, and  furthermore, the corresponding losses converge to zero.
	Our approach to establish (i)  is based on the adaptation in our current probabilistic setting  of the liminf-limsup framework of De Giorgi, see Section 2.3.4 of \cite{DeGiorgi_sel_papers:2013} for an historical account, commonly used in the $\Gamma-$convergence of functionals in the calulus of variations. The results on local minimisers hinges on a modification of the analysis for the global minimisers, but restricting our attention only to the liminf bounds. Finally the convergence result  of $(u_{\bm {X}_{N(\ell)} (\omega)} )_\ell$      to $f,$ as both the network richness and the number of samples grow is based on an  {$L_1$-estimate of the loss function.}
	These results show that generic neural network algorithms considered in this work have the right asymptotic behaviour, under no extra assumptions. This can be seen as a demonstration of the robustness of the algorithms,   when they are based on the existence of a well defined underlined target function, regardless of its smoothing properties. 
	Notably, our analysis provides useful information on the behaviour of local minimisers as well, under, in principle, verifiable conditions in practical computations.   Thus,  this analysis, combined with the estimates provided in Section 4,  provides a rather clear picture of the asymptotic behaviour of learning algorithms, and in addition,  opens new possibilities on the development of improved tools and algorithms.

 \subsubsection*{Remarks on bibliography} 	
	In this work we avoid structural hypotheses on the neural network spaces. 
	Clearly, in order to estimate the approximation errors due to DNN, see Theorem \ref{Thm:estimate}, and assumption  \eqref {w_ell_7},  one has to use known approximability  estimates. There is vast activity over the last years devoted to the approximation capability of DNN spaces, see for example, 
	\cite{Dahmen_Grohs_DeVore:specialissueDNN:2022,Schwab_DNN_constr_approx:2022, Xu_appr_cnn:2022, Xu_approx:2022, Schwab_DNN_highD_analystic:2023, Mishra:appr:rough:2022, Grohs_Petersen_Review:2023} and their references. New results are needed 
	among other topics, on specifying architectures required for given  approximation bounds, instead of typical existence results. Nevertheless, 
	the remarkable advancements in the approximation theory of neural networks
	constitute a milestone in understanding their behaviour. \\
	As mentioned, Rademacher complexity tools, are among the most popular methods 
	to estimate generalisation error. Related tools include,   VC dimension, local Rademacher complexity, kernel methods, see for example, 
\cite{Oneto_local_Rademacher:2018}  
\cite{Yousefi_local_Rademacher:2018}
\cite{Bartlett_VCdimension:2019}
\cite{Jegelka_gen_opt_transport:2021}
\cite{Bartlett_Benignoverfitting:2020}
\cite{WE_Rademacher:2020}
\cite{Montanari_neural_kernel:2021}
\cite{chenVanden-Eijnden_feature:2022} and their references. \\ 
Optimal transport framework has emerged as an important tool in data science and machine learning, see for example, \cite{Peyre:CompOT:2019}, \cite{PanaretosWR:2019},
and their references. More recent works include \cite{Jegelka_gen_opt_transport:2021}, \cite{Kats_Pant_2022optimizing}, 
\cite{Peyre2023unbalanced}.
In the work of \cite{Jegelka_gen_opt_transport:2021} the  Rademacher complexity of a class of neural network spaces is controlled using Wasserstein distances
and their relation to correlation. Still it is assumed that the members of the class satisfy  certain structural hypotheses, and it is demonstrated that
such bounds provide useful empirical tools to control the generalisation error. \\
Part of the convergence analysis of section 5 is motivated by typical $\Gamma$-convergence arguments in the Calculus of Variations. $\Gamma$-convergence is a very natural and nonlinear approach which is extensively used in nonlinear energy minimisation in various instances.  
See \cite{muller2020deep} for an application to deep Ritz methods without training. Recently, 
 this approach motivated the introduction of suitable stability notions and corresponding convergence analysis for a class of 
neural network problems approximating partial differential equations, \cite{GGM_pinn_2023}. 
In our work, given the probabilistic character of the model and the separation of the sources of discretisation (training samples and discrete DNN spaces), the application of this framework 
 is done in a 
non-standard  setting. 
In the analysis of Section 5, we still use the representation of the loss through empirical measures and a crucial result related to relative compactness of measures from \cite{AGS:2005}.


\section{Probabilistic setting  and preliminaries}
We consider a collection $Y, X_1,X_2,\ldots$ of i.i.d.\ $D$-valued random variables, defined on a probability space $(\Omega,\mathcal{F},\PP)$, with common law $\mu.$ 
For $g\in L_p(D,\mu)$ we have 
\begin{equation}\label{repr_one}
	\int _D \, g (x) \, \d \mu (x) = \int _\varOmega \, g ( X_i (\omega)) \, \d \PP (\omega) = \int _\varOmega \, g ( Y (\omega)) \, \d \PP (\omega)= \EE [ \, g(Y) \, ]
\end{equation} 
In particular, 
\[
\|u_{\x}-f\|_p^p=\int_D|u_{\x} (x)-f(x)|^p\, \d \mu(x)=\EE\big[|u_{\x} (Y)-f(Y)|^p\big]\, . 
\]
 To analyse \eqref{ieE-minimize_NNQ} we need to reformulate  it in a suitable probabilistic framework. To this end, 
let $\omega \in \varOmega $ be fixed. Consider the discrete energy, 
\begin{equation}\label{prob_E}
\mathcal{E}_{N, \omega}( g )  =  \frac 1N  \sum _{i}   \Big  | g(X_i(\omega)\, ) - f(X_i (\omega) \, ) \Big  | ^p  \, . 
\end{equation}

\begin{definition} \label{D-dnnint}{\sc [probabilistic deep neural network interpolation]}  
Assume that for each $\omega \in \varOmega$ the problem 
\begin{equation}\label{prob_E_def}
	\min  _ {v \in  V _{\mathcal{N}} } \E _{N, \omega} (v)
\end{equation}
has a solution $u ^\star(\omega, \cdot)   \in V _{\mathcal{N}} .$ We denote by   $ u_{\bm{X} (\omega)   }  = u ^\star(\omega, \cdot)  \,  $ the  $V_{\mathcal{N}}-$\emph{ probabilistic deep neural network interpolant of $f$ at $ \bm{X}  = (X_1, \dots, X_N  )  \, .$}
\end{definition}

\subsection{Formulation and  remarks.} A crucial tool in the subsequent analysis is to express the discrete loss and other quantities in terms of \emph{empirical measures.}\,  Then, through the optimal transport framework, we will estimate distances between measures, and
study their weak convergence; hereafter, we use the notation of \cite {AGS:2005}.  The following setting and observations will be useful in the sequel.\\

\subsubsection{Representation through empirical measures.}
For fixed $\omega$ consider the empirical measure on $D,$
\begin{equation}\label{emp_meas}
	\mu _{N, \bm{X} (\omega)} =   \frac 1 N \sum _{i} \   \delta _{X_i(\omega)} \, ,
\end{equation}
Then, the discrete energy can be written as
\begin{equation}\label{energy_emp_meas}
	\mathcal{E}_{N, \omega}( g )  = \int _D |g (x)-f(x)|^p\, \d \mu _{N, \bm{X} (\omega)} (x) 
\end{equation}
Furthermore, 
\begin{equation}\label{energy_emp_meas0}
\EE\big[ \mathcal{E}_{N, \omega}( g ) \big] 
	=\EE \big[  \frac 1N  \sum _{i=1}^N   \Big  | g(X_i \, ) - f(X_i   \, ) \Big  | ^p \big] =   \frac 1N  \sum _{i=1}^N  \EE \Big[  \,   \big  | g(X_i \, ) - f(X_i   \, ) \big  | ^p \Big] =\|g-f\|_p^p.
\end{equation}

\subsubsection{Distance of measures.}  We introduce   the Wasserstein distance between measures on $D$ and express the various quantities of interest via the optimal transport framework. Let $p \geq 1 $ and $\mu, \nu$ probability measures on $D$   with finite $p$ moments.  The Wasserstein distance of order $p$  between $\mu, \nu$ is defined by 
\begin{equation}\label {W-dist}
	W_p(\mu, \nu) = \Big \{ \min_{\pi \in A (\mu, \nu)} \int _{D\times D} |x-y|^p\, \d \pi(x,y) \Big \} ^{1/p} ,
\end{equation}
where $A (\mu, \nu)$  is the set of plans  between $\mu$ and $ \nu,$ i.e.,   Borel probability measures  on $D \times D$  whose first and second marginals
are $ \mu$ and $ \nu$, respectively. For $\mu $ a probability measure on $D$, and $ \xi  : D \to \R $ a  Borel measurable map, we denote by $\xi _\sharp \mu $  the push-forward of %
$\mu $ through $\xi$, i.e., the probability measure on $\R $  defined by 
$$ \xi _\sharp  \mu (B) :=  \mu(\xi ^{-1} (B)),\qquad \text{Borel } B\subset \R\, . $$ 
In particular, 
\begin{equation}\label {pf-int}
	 \int _{D } |\xi (x)|^p\, \d \mu(x)  =  \int  |s|^p\, \d \xi _\sharp \mu (s)  .
\end{equation}
With the obvious modification in the definition of $W_p$, 
we have, 
\begin{equation}\label {W-dist}
	W_p(\xi _\sharp  \mu, \xi _\sharp \nu) = \Big \{ \min_{\pi \in A (\mu, \nu)} \int _{D\times D} |\xi (x) -\xi (y) |^p\,  \d \pi(x,y) \Big \} ^{1/p} \, .
\end{equation}
Without loss of generality we assume that $0\in D$ and denote by $\delta_0 $ the Dirac probability  measure concentrated at $0.$
Then, 
we have $ A (\mu, \delta_0)= \{ \, \mu \times \delta_0 \, \}  $
\cite[(5.2.12)]{AGS:2005}, and
\begin{equation}\label {W-dist-d_0}
	W_p(\mu, \delta_0) = \Big \{ \min_{\pi \in A (\mu, \delta_0)} \int _{D\times D} |x-y|^p\, \d \pi(x,y) \Big \} ^{1/p} 
=\Big \{ \  \int _{D } |x|^p\, \d \mu(x) \Big \} ^{1/p}\, .
\end{equation}
\subsubsection{Bound of discrete energies.} By definition we have for all $v \in  V_{\mathcal{N}},$ and all $\omega \in \varOmega\, , $
\begin{equation}\label{prob_E_min_prop}
  \frac 1N  \sum _{i}   \Big  | u_{\bm{X} (\omega)}(X_i(\omega)\, ) - f(X_i (\omega) \, ) \Big  | ^p  \leq 
\frac 1N  \sum _{i}   \Big  | v (X_i(\omega)\, ) - f(X_i (\omega) \, ) \Big  | ^p  \, . 
\end{equation}
By taking expectations of both sides and using \eqref{energy_emp_meas0} we get
\begin{equation}
\label{prob_E_min_prop3}
\EE \Big[ \frac 1N  \sum _{i=1}^N   \big  | u_{\bm{X}  }(X_i \, ) - f(X_i   \, ) \big  | ^p \Big]
 \leq  \|v-f\|_p^p
 ,\qquad\text{for all } v \in  V_{\mathcal{N}}.
\end{equation}

\section{Estimates through optimal matching}
\label{sec:Estimates_matching}
Our goal in this section is to reduce the problem of estimating the 
error 
\begin{equation}\label {W-dist-d_0}
\EE\big[ \|u_{\bm {X}(\cdot)}-f\|_p^p\big]\ 
=\int _\Omega \int_D|u_{\bm {X}(\omega)} (x)-f(x)|^p\, \d \mu(x) \d \P (\omega)\, .
\end{equation}
to a simple estimate involving two terms: the first one is an pure 
approximability term reflecting the approximation capacity of the $V_{\mathcal{N}}$ space and the second term is a term reflecting the error of Monte-Carlo integration in this non-trivial setting. The second term is directly connected to 
\emph{optimal matching problems} in probability. 
 These are  random problems in a quite active area 
 of probability, mathematical physics and their applications related to estimating appropriate distances of empirical measures to a common law, or distances of two sets of randomly selected points in domains of $\R^d,$ see  \cite{FournierGuillin:2015}\cite{AjtaiKT:1984}\cite{AmbrosioST:2019,AGS:2005,AmbrosioGT:2019,AmbrosioGT:2022,Parisi:matching:2014,Talagrand:1994,Talagrand:1992,Talagrand:AKTgeneral:1992,Talagrand:book}, see also \cite{PanaretosWR:2019} for the relevance of these bounds in statistics. 
 We have the following
 \begin{Theorem}[\sc estimate of the generalisation error]
\label{Thm:estimate}
Consider for each $\omega \in \varOmega$ the problem 
\begin{equation}\label{prob_E_def}
	\min  _ {v \in  V _{\mathcal{N}} } \E _{N, \omega} (v)
\end{equation}
and its solution     $ u_{\bm{X} (\omega)   },    $ the  $V_{\mathcal{N}}-$\emph{probabilistic deep neural network interpolant of $f$ at $ \bm{X}  = (X_1, \dots, X_N  )  \, .$}
Assume that $f$ is Lipschitz and let us denote the Lipschitz constant of $ u_{\bm{X} (\omega)   }  -f$ by $L_{\bm{X} (\omega)   }\, .$ Then for each $\omega\in \varOmega, $ and
$\varphi   \in  V _{\mathcal{N}},$
\begin{equation}\label {W-dist-d_0}
  \|u_{\bm {X}(\omega)}-f\|_p \leq \Big \{  \frac 1N  \sum _{i}   \Big  | \varphi  (X_i(\omega)\, ) - f(X_i (\omega) \, ) \Big  | ^p  \Big \}^{1/p} + 
L_{\bm{X}}\,   W_p(\mu, \mu _{N, \bm{X} (\omega)}) \, .
\end{equation}
Furthermore, if 
\begin{equation}\label {W-dist-d_0}
L_{\bm{X}}\,  \leq L_N,\quad \mathbb{P}-a.s.
\end{equation}
we have
\begin{equation}\label {W-dist-d_0}
 \EE\big[ \|u_{\bm {X}(\cdot)}-f\|_p \big]  \leq  \inf_{\varphi   \in  V _{\mathcal{N}}}  \|\varphi-f\|_p
  + 
L_N\,  \EE\big[  W_p(\mu, \mu _{N, \bm{X} (\cdot )})  \big]\, .
\end{equation}
\end{Theorem}

\begin{proof}
Notice that 
\begin{equation}\label {W-dist-d_0_2}
\begin{split}
\Big \{ \  \int _{D } \big |\big [u_{\bm {X}(\omega)}-f\, \big ](y)\big|^p\, \d \mu(y) \Big \} ^{1/p} =	W_p(\big [u_{\bm {X}(\omega)}-f\, \big ]_\sharp\mu, \delta_0) \, .
\end{split}
\end{equation}
Then, since $W_p$ is a distance and, $u_{\bm {X}(\omega)}$ is a minimiser of the discrete problem, 
\begin{equation*}\label {W-dist-d_0_3}
\begin{split}
\Big \{ \  \int _{D } \big |\big [u_{\bm {X}(\omega)}-f\, \big ](y)\big|^p\, \d \mu(y) \Big \} ^{1/p}
 \leq	&W_p(\big [u_{\bm {X}(\omega)}-f\, \big ]_\sharp\mu _{N, \bm{X} (\omega)}, \delta_0)\\
  &+  W_p(\big [u_{\bm {X}(\omega)}-f\, \big ]_\sharp\mu, \big [u_{\bm {X}(\omega)}-f\, \big ]_\sharp\mu _{N, \bm{X} (\omega)})\\
 = &\Big \{  \frac 1N  \sum _{i}   \Big  | u_{\bm {X}(\omega)} (X_i(\omega)\, ) - f(X_i (\omega) \, ) \Big  | ^p  \Big \}^{1/p}\\
 &+  W_p(\big [u_{\bm {X}(\omega)}-f\, \big ]_\sharp\mu, \big [u_{\bm {X}(\omega)}-f\, \big ]_\sharp\mu _{N, \bm{X} (\omega)})\\
 \leq &\Big \{  \frac 1N  \sum _{i}   \Big  | \varphi  (X_i(\omega)\, ) - f(X_i (\omega) \, ) \Big  | ^p  \Big \}^{1/p}\\
 &+  W_p(\big [u_{\bm {X}(\omega)}-f\, \big ]_\sharp\mu, \big [u_{\bm {X}(\omega)}-f\, \big ]_\sharp\mu _{N, \bm{X} (\omega)})
  \, .
\end{split}
\end{equation*}
The first assertion then follows by observing, 
\begin{equation*}\label {W-dist-d_0_4}
\begin{split}
  W_p(\big [ u_{\bm {X}(\omega)}-&f\, \big ]_\sharp\mu, \big [u_{\bm {X}(\omega)}-f\, \big ]_\sharp\mu _{N, \bm{X} (\omega)}) \\
  =\, & \Big \{ \min_{\pi \in A (\mu, \mu _{N, \bm{X} (\omega)})} \int _{D\times D} |\big [ u_{\bm {X}(\omega)}- f\, \big ] (x) -\big [ u_{\bm {X}(\omega)}-f\, \big ] (y) |^p\,  \d \pi(x,y) \Big \} ^{1/p} \\
  \leq \, & L_{\bm{X} (\omega)   } \, W_p( \mu, \mu _{N, \bm{X} (\omega)}) 
  \, .
\end{split}
\end{equation*}
As mentioned, 
\begin{equation*}\label{prob_E_min_prop2}\begin{split} 
\int _\varOmega \,  \frac 1N  \sum _{i}   \Big  | \varphi  (X_i(\omega)\, ) - f(X_i (\omega) \, ) \Big  | ^p   \, \d \PP (\omega) \\
  = \int_D|\varphi(x)-f(x)|^p\, \d \mu(x)\, . 	
\end{split}
\end{equation*}
and the proof is complete.
\end{proof}
\begin{remark}[\sc a priori control of the empirical error]
The previous proof shows that the error induced by Monte-Carlo sampling is
essentially controlled by
$$ W_p(\big [u_{\bm {X}(\omega)}-f\, \big ]_\sharp\mu, \big [u_{\bm {X}(\omega)}-f\, \big ]_\sharp\mu _{N, \bm{X} (\omega)}) \, .$$
Instead of just controlling $L_{\bm{X}}\,   W_p(\mu, \mu _{N, \bm{X} (\omega)})$ it would be interesting  to estimate directly the above term. 
This is an interesting  problem, which opens alternative possibilities for finer control of the generalisation error.
	
\end{remark}
The problem of asymptotic behaviour of $ \EE [   W_p(\mu, \mu _{N, \bm{X} (\omega)})]$ has been studied from various perspectives over the years. 
More recently, PDE techniques were proven to be useful in deriving upper and lower bounds, see  \cite{AmbrosioST:2019, AmbrosioGT:2019,AmbrosioGT:2022}. 
The results so far suggest that 
\begin{equation}\label{emp_W_rates0}
	 \EE\big[  W_p(\mu, \mu _{N, \bm{X} (\cdot )})  \big]  \asymp N^{-p/d}\, ,
\end{equation}
a behaviour that has been conjectured in the general case in \cite{Parisi:matching:2014}. Notice that in the critical $d=2$ case \eqref {emp_W_rates0} holds up to a logarithmic factor. 
As far as the upper bounds are concerned, the results of \cite{FournierGuillin:2015}, imply 
\begin{equation}\label{emp_W_rates}
	 \EE\big[  W_p(\mu, \mu _{N, \bm{X} (\cdot )})  \big]  \leq C(\mu)\,  N^{-p/d}\, ,
\end{equation}
provided that $\mu $ has bounded $p-$moments, and $ C(\mu)$ includes a logarithmic term for $d=2.$ We thus have:
\begin{Theorem}[\sc asymptotic bound of the generalisation error]
\label{Thm:estimate}
Let    $ u_{\bm{X} (\omega)   }    $ be the  $V_{\mathcal{N}}-$\emph{probabilistic deep neural network interpolant of $f$ at $ \bm{X}  = (X_1, \dots, X_N  )  \, .$}
Assume that $f$ is Lipschitz and let us denote the Lipschitz constant of $ u_{\bm{X} (\omega)   }  -f$ by $L_{\bm{X} (\omega)   }\, .$ If \begin{equation}\label {W-dist-ass}
L_{\bm{X}}\,  \leq L_N
\end{equation}
we have
\begin{equation}\label {W-dist-d_0}
 \EE\big[ \|u_{\bm {X}(\cdot)}-f\|_p \big]  \leq  \inf_{\varphi   \in  V _{\mathcal{N}}}  \|\varphi-f\|_p
  + 
L_N\, C(\mu)\,  N^{-p/d} \, .
\end{equation}
Furthermore, if 
\begin{equation}\label {W-dist-d_0}
   \inf_{\varphi   \in  V _{\mathcal{N}}}  \|\varphi-f\|_p \to 0 \quad \text{and} \quad
L_N\,   N^{-p/d} \to 0 \, , 
\end{equation}
as $|\mathcal{N}| \to \infty $ and $N \to \infty$ respectively
we conclude that 
\begin{equation}\label {W-dist-d_0}
 \EE\big[ \|u_{\bm {X}(\cdot)}-f\|_p \big] \to 0\, .
\end{equation}
\end{Theorem}

\bigskip

\begin{remark}[\sc dimension dependence] The above bounds do not depend favourably 
on the dimension $d$, since the rates are deteriorating in high dimensions.   
However, new dimension dependent results, \cite{hundrieser2022empirical},	suggest that under further assumptions on the measure $\mu,$ (e.g.
if $\mu$ is concentrated on a lower dimensional manifold) these rates can be improved significantly.
\end{remark}

\begin{remark}[\sc structural assumptions] Although we do not assume any structural assumptions on the spaces $ V _{\mathcal{N}},$ the bounds remain obviously valid for spaces 
that are designed to preserve, for example,  uniform bounds of the Lipschitz constant. 
\end{remark}

\section{Convergence results}
\subsection{The $N\to \infty$ limit}\label{sec:compactness_convergence}

\bigskip

We now fix the DNN architecture and the corresponding space $V _{\mathcal{N}}$, and we take to infinity the number of samples, $N$. We will prove that, with probability 1, the family of minimisers has weak limits whose generalisation error satisfies an optimal approximation error bound.  A key tool in our approach is De Giorgi's liminf - limsup framework, commonly used in the  $\Gamma$-convergence of functionals, adapted in our present probabilistic setting.\\
Our first goal  is to show for a.e.\  $\omega \in   \varOmega $ the existence of a function $u_\omega^\mathcal{N} \in L^p (D, \mu)$ such that 
\begin{align}
 \E (u_\omega^\mathcal{N}) \leq \liminf\limits_{N\rightarrow \infty}  \E _{N, \omega}  [ u_{\bm {X}_{N}(\omega)} ]\, ,
\end{align}
and $u_{\bm {X}_{N} (\omega)} \to u_\omega^\mathcal{N}$ in an appropriate sense. \\
We recall first the notion of weak convergence of measures: A sequence of measures $\nu_n $ on $D$ \emph{narrowly (weakly) converges} to $\nu$ if for all bounded and continuous functions, $g : D \to \R$ we have 
 \begin{equation}\label{narrow-conv0}\begin{split} 
\lim _{n\to \infty} \int _D \,  g (x) \, \d \nu_n (x) = \int _D \,  g (x) \, \d \nu (x) \, . 	
\end{split}
\end{equation}
If $\nu_n $ weakly converges to $\nu$, we say  \cite[Definition 5.4.3]{AGS:2005}\, that a sequence of functions $v_n \in L^1 (D, \nu_n)$ weakly converges to 
$v  \in L^1 (D, \nu)$ with respect to $(\nu_n, \nu), $ if 
 \begin{equation}\label{w-conv-u_n0}\begin{split} 
\lim _{n\to \infty} \int _D \, \zeta (x) v_n(x) \, \d \nu_n (x) = \int _D \,  \zeta (x) v(x)  \, \d \nu (x) \, , \qquad\text{for all }\zeta \in C_0^\infty (D)\, .
\end{split}
\end{equation}
We also say that $v_n$ strongly converges to $v$ in $L_p,\ p > 1$, with respect to $(\nu_n, \nu), $ if \eqref{w-conv-u_n0} holds and
\begin{equation}\label{strongconv}
\limsup _{n\to \infty}\|v_n\|_{L_p(D,\nu_n)} \leq \|v\|_{L_p(D,\nu)} .
\end{equation}\\
Let us also recall the following multidimensional generalisation of the Glivenko-Cantelli theorem.
\begin{Lemma}\label{GC}
Let $X_1,X_2,\ldots$ be a sequence of i.i.d. random variables with common law $\mu$. There exists a subset $\varOmega _1\subset \varOmega $ with $\P ( \varOmega \backslash  \varOmega _1)=0$, such that 
\begin{equation}\label{perklln}
 \mu _{N, \bm{X}_N (\omega  )} \to \mu, \qquad \text{ for all }\omega \in \varOmega _1.
 \end{equation}
\end{Lemma}
A proof of Lemma \ref{GC} can be found in \cite{V-58}, and follows from the fact that there exists a fixed countable subset $\mathcal {G}$ of the set of bounded continuous functions on $D$, for which it suffices to check the validity of \eqref{narrow-conv0} to conclude weak convergence of measures. For each such $g\in \mathcal {G},$ we have 
\begin{equation*} \begin{split}
\int g(x)\, \d \mu _{N, \bm{X}_N (\omega  )}(x)= \frac 1N  \sum _{i=1}^N    g (X_i)    \,   \longrightarrow  \int_D g(x)\, \d \mu(x)\, , \qquad \mathbb{P}-\text{a.s.}, 	
\end{split}
\end{equation*}
by the strong law of large numbers, cf. \cite{Kallenberg_book}. Since any countable union of sets of measure zero has measure zero, there exists a set $\varOmega _1$, such that \eqref{perklln} holds. \\

\begin{Proposition}[\sc the $\liminf$ inequality]
\label{Proposition:liminf_N}
Let $p>1$ and $V _{\mathcal{N}}$  a fixed DNN space. For $\omega\in \varOmega$, consider a sequence of  absolute minimisers of 
\begin{equation}\label{prob_E_def_theorem0}
	\min  _ {v \in V _{\mathcal{N}} } \E _{ {N } , \omega} (v)
\end{equation}
denoted by   $ u_{\bm{X}_{N} (\omega)   }  ,$ where $ \bm{X} _{N} = (X_1, \dots, X_{N}  )  \, .$
Assume that there exists a  subset $\varOmega _2\subset \varOmega   $ with $\P ( \varOmega \backslash  \varOmega _2)=0$, such that
\begin{equation}\label{w-conv-u_n_bound0}\begin{split} 
   \E _{N, \omega}  \big( u_{\bm {X}_N (\omega)} \big) \leq M (\omega)<+\infty\, ,
   \qquad \text{ for all  }\ \omega \in \varOmega _2.
\end{split}
\end{equation}	
Then, for $\mathbb{P}$-a.e. $\omega\in\Omega$, the sequence $(u_{\bm{X}_{N} (\omega)   } )_N$  has weak limits along subsequences with respect to $ (\mu _{N, \bm{X}_{N} (\omega  )}, \mu) $, and for any such weak limit, $ u _{\omega} ^  {\mathcal{N}}$, we have
\begin{equation}\label{w-conv-u_n_0}\begin{split} 
\E \big(u_\omega^  {\mathcal{N}}\big) \leq \liminf\limits_{N\rightarrow \infty}  \E _{{N}, \omega}  \big( u_{\bm {X}_{N} (\omega)} \big)\, .
\end{split}
\end{equation}	
\end{Proposition}
\begin{proof}
The proof is an adaptation to our present setting of \cite[Theorem 5.4.4]{AGS:2005}. 
Assumption \eqref{w-conv-u_n_bound0} implies that 
\begin{equation}\label{energy_emp_meas0}
	 \int _D |u_{\bm {X}_N (\omega)} (x)-f(x)|^p\, \d \mu _{N, \bm{X}_N (\omega)} (x)=\E _{N, \omega}  \big( u_{\bm {X}_N (\omega)} \big) \leq M (\omega ),  \quad \omega \in \varOmega _2.
\end{equation}
This implies that the sequence of measures $\big(\big[u_{\bm {X}_N (\omega)} -  f \big]_\sharp \mu _{N, \bm{X}_N (\omega)} \big)_N=\big([{v_N (\omega)}]_\sharp \mu _{N, \bm{X}_N (\omega)}\big)_N $ is relatively compact. It is not obvious however that the limit points of this sequence of measures will be of the form $ [{v (\omega)}]_ \sharp \mu \, .$ 
To this end, following \cite[Theorem 5.4.4]{AGS:2005}, consider $\omega\in\varOmega_1\cap\varOmega_2$, and define the plans  $\bm{\gamma} _N  = (\bm{i} \times  {v} _N )_\sharp \mu _{N, \bm{X}_N }, $  where
$\bm{i}: D\to D$ is the identity map. The first marginal of $\bm{\gamma} _N $ is 
$\pi ^1_\sharp \bm{\gamma} _N = \mu _{N, \bm{X}_N} $ and the second, 
$\pi ^2 _\sharp \bm{\gamma} _N =  ({v} _N )_\sharp  \mu _{N, \bm{X}_N }.$ Since $\pi ^1 _\sharp \bm{\gamma} _N= \mu _{N, \bm{X}_N} $ weakly converges to $\mu$ for $\omega\in\varOmega_1$, and $\pi ^2 _\sharp \bm{\gamma} _N $ is relatively compact for $\omega\in\varOmega_2$, 
it follows from \cite[Theorem 5.4.4]{AGS:2005} that $\bm{\gamma} _N$ is relatively compact on $\varOmega_1\cap\varOmega_2$. For the rest of the proof we will assume that $\omega\in\varOmega_1\cap\varOmega_2$, a set a probability 1, and we will not explicitly show dependence on $\omega$. Let $\bm{\gamma} $ be a weak subsequential limit 
of $\bm{\gamma} _{N} $. Since $p>1$ and 
\[
\int |y|^p \d\bm{\gamma} _N(x,y)=\int_D |u_{\bm {X}_N} (x)-f(x)|^p\, \d \mu _{N, \bm{X}_N } (x) < +\infty,
\]
we conclude that $|y | $ is $ \bm{\gamma} _{N}$ uniformly integrable. Hence, along the subsequence for which $\bm{\gamma}_N\to \bm{\gamma} $ we have
for all $\zeta \in C_0^\infty (D)\, ,$
\begin{equation}\label{w-conv-u_n_00}\begin{split} 
\lim _{N\to \infty} \int _D  \, \zeta (x) \,v_{N} (x) \d \mu _{ {N}, \bm{X}_{N}} (x) 
&= \lim _{N\to \infty} \int _{D \times \R}   \zeta (x)\,  y \,   \d \bm{\gamma} _{N} (x, y) \\
&= \int _{D \times \R}   \zeta (x)\,  y \,   \d \bm{\gamma}   (x, y)  \, , \\ 
\end{split}
\end{equation}
Notice however that the first marginal of $ \bm{\gamma} $ is $\mu $, thus 
\begin{equation}\label{w-conv-u_n_03}\begin{split} 
 \int _{D \times \R}   \zeta (x)\,  y \,   \d \bm{\gamma}   (x, y) 
= \int _{D}   \zeta (x)\,     \overline \gamma_ 1(x)    \d  \mu  (x)  \, , \\ 
\end{split}
\end{equation}
where $  \overline {\bm{\gamma}}_1  (x)  = \int _{\R }\, y\,  \d \bm{\gamma}_{x} (y)$ is the barycentric projection with respect to the first marginal of $\bm {\gamma}, $  \cite[p.\ 126-8]{AGS:2005}. 
Relationship \eqref{w-conv-u_n_0} implies that $  \overline \gamma_ 1$ in general depends on $\omega .$ We thus set $ u_\omega^{\mathcal{N}} (x) =    \overline \gamma_ 1 (x) +f (x)\, .$ 
Passing to the limit along a subsequence such that $\bm{\gamma} _{N}   \to \bm{\gamma}  $ we conclude that
\begin{equation}\label{w-conv-u_n_4}\begin{split} 
 \ \liminf\limits_{N\rightarrow \infty}   \E _{{N}, \omega}  \big( u_{\bm {X}_{N} (\omega)} \big) &=
  \ \liminf\limits_{N\rightarrow \infty}  \int _{D \times \R}   \, | y | ^p\,   \d \bm{\gamma} _{N  } (x, y) \geq
   \int _{D\times \R}    | y | ^p \,   \d \bm{\gamma}   (x, y)  \, , \\ 
   &=\int_D \Big( \int_{\R}\, |y|^p \d\bm{\gamma}_x(y)\Big)\d \mu(x)\\ 
   &\geq \int _{D}       |  \overline \gamma_ 1(x)  |^p  \d  \mu  (x) = \E \big(u_\omega^{\mathcal{N}}\big) \, ,
\end{split}
\end{equation}
and the proof is complete. 
\end{proof} \\

We now establish an upper bound for the discrete energy of the minimisers in terms of the error in the approximation of $f$ by elements of the network.
\begin{Proposition}[\sc the $\limsup$ inequality]
\label{Thm:limsup_N}
Let  $V _{\mathcal{N}}$ be a fixed DNN space. For $\omega\in \varOmega$ consider a sequence $ \big(u_{\bm{X}_{N} (\omega) }\big)_N    ,$ of minimisers of 
\eqref{prob_E_def_theorem0}. Then, 
\begin{equation}\label{limsupbound}
\limsup_{N\to\infty}\E _{ {N } , \omega} (u_{\bm{X}_{N } (\omega)   }  )\leq  \inf_{\varphi   \in  V _{\mathcal{N}} } \|\varphi-f\|_p^p\, ,\qquad\text{for}\ \mathbb{P}-\text{a.e. } \omega\in\varOmega.
\end{equation}
\end{Proposition}

\begin{proof} 
The minimising property of $u_{\bm{X}_{N} (\omega)}$ implies that
\begin{equation}\label{prob_E_theorem_2_10}
	\E _{ {N } , \omega} (u_{\bm{X}_{N } (\omega)   }  ) \leq \E _{ {N } , \omega} (v)\, , \quad \text {for all } \ v \in  V _{\mathcal{N}}  \, . 
\end{equation}
Next, take a sequence $(v_n)_n$ in $V _{\mathcal{N}}$ that realises the infimum in \eqref{limsupbound}, i.e.,
\[
\lim_{n\to\infty} \|v_n-f\|_p=\inf_{\varphi   \in  V _{\mathcal{N}} } \|\varphi-f\|_p.
\]
For each $v_n$  take $N\to\infty$ in \eqref{prob_E_theorem_2_10}. By the law of large numbers, there exists a set  $E_n\subset\varOmega$, with $\mathbb{P}\big[E_n\big]=1$, such that for $\omega\in E_n$,
\begin{equation}\label{llnvn}
\lim_{N\to\infty} \E _{ {N } , \omega} (v_n)=\lim_{N\to\infty} \frac 1N \sum_{i=1}^N |v_n(X_i)-f(X_i)|^p=\|v_n-f\|_p^p.
\end{equation}
Let us now set $E=\cap_n E_n$. Then, $\mathbb{P}\big[E\big]=1$ and, in view of  \eqref{prob_E_theorem_2_10} and \eqref{llnvn}, for all $\omega\in E$ we have
\begin{equation}\label{w-prob_E_theorem_3}
\limsup_{N\rightarrow \infty}   \E _{N, \omega}  ( u_{\bm {X}_{N} (\omega)} )   
 \leq \inf_n\|v_n-f\|_p^p =\inf_{\varphi\in V_\mathcal{N}} \|\varphi-f\|_p^p  \, ,
\end{equation}	
concluding the proof of the proposition.
\end{proof}
  We may thus combine the results of the two preceding propositions into the following theorem.\\

\begin{Theorem}[\sc limit behaviour of minimisers as $N\to \infty$]\label{Ntoinftythm}
Let $p>1$ and $V _{\mathcal{N}}$  a fixed DNN space. For $\omega\in \varOmega$, consider a sequence $ \big(u_{\bm{X}_{N} (\omega)   } \big)_N$  of  absolute minimisers of 
\[
	\min  _ {v \in V _{\mathcal{N}} } \E _{ {N } , \omega} (v),
\]
where $ \bm{X} _{N} = (X_1, \dots, X_{N}  )  \, .$
Then, for $\mathbb{P}$-a.e.\ $\omega\in\Omega$, the sequence $(u_{\bm{X}_{N} (\omega)   } )_N$  has weak limits along subsequences with respect to $ (\mu _{N, \bm{X}_{N} (\omega  )}, \mu) $, and for any such weak limit, $ u _{\omega} ^  {\mathcal{N}}$, we have
\begin{equation}\label {est_final}
  \|u _{\omega} ^  {\mathcal{N} } -f\|_p   \leq  \inf_{\varphi   \in  V _{\mathcal{N}}}  \|\varphi-f\|_p
 \, .
\end{equation}
\end{Theorem}

\begin{proof} 
Note that \eqref{limsupbound}, in particular implies assumption \eqref{w-conv-u_n_bound0} in Proposition \ref{Proposition:liminf_N}.
Therefore,   we conclude that 
there exists a    $u_\omega  ^  {\mathcal{N} } \in L^p (D, \mu)$ such that $u_{\bm {X}_{N }(\omega)} $ converges weakly {up to a subsequence} to $u_\omega ^  {\mathcal{N} }$ with respect to $ (\mu _{\ell, \bm{X}_{N(\ell)} (\omega  )}, \mu) $ and  
\begin{equation}\label{w-prob_E_theorem_2}\begin{split} 
\E (u_\omega ^  {\mathcal{N} })  \leq \liminf\limits_{N \rightarrow \infty}  \E _{N, \omega}  [ u_{\bm {X}_{N} (\omega)} ]\, .
\end{split}
\end{equation}	
Then we finally obtain, 
\begin{equation}\label{w-conv-u_n_final}
\E \big(u_\omega^  {\mathcal{N}}\big) \leq \liminf\limits_{N\rightarrow \infty}  \E _{{N}, \omega}  \big( u_{\bm {X}_{N} (\omega)} \big)\,\leq \limsup_{N\rightarrow \infty}   \E _{N, \omega}  ( u_{\bm {X}_{N} (\omega)} )   
 \leq \inf_{\varphi\in V_\mathcal{N}} \|\varphi-f\|_p^p  \, .
\end{equation}	
and the proof is complete.
\end{proof}

\begin{remark}[\sc robustness for large $N$] One of the interpretations of the above bounds is that the learning algorithm is robust with respect to large values of samples. 
In fact, all possible weak limits of the sequence, will have an $L^p$ distance from $f$ which is only controlled by the optimal approximation error  of the space $V_\mathcal{N}\, .$ 
Hence, if the number of samples is large enough, we expect that the dominant source of the error will be the DNN approximation error. 
\end{remark}

\subsection{Asymptotic behaviour of local minimisers as $N\to \infty$}\label{subsec:local_min}

\bigskip

Next, we keep fixed the DNN architecture and the corresponding space $V _{\mathcal{N}}$, and we study the behaviour of \emph{local minimisers} as  we take to infinity the number of samples, $N$. Local minimisers can be reached, in certain cases, by approximate methods to the minimisation problem, for example when stochastic gradient is used. Typically, in these situations we know the corresponding loss, but it is not clear how the generalisation error behaves. The  results of this section assume certain properties for the loss at local minima. We will prove that, if the losses of a given family of local minimisers are bounded then still,  
 for a.e.\  $\omega \in   \varOmega ,$ there exists  a function $u_{\omega, \text{loc}}^\mathcal{N} \in L^p (D, \mu)$ such that 
\begin{align}
 \E (u_{\omega, \text{loc}}^\mathcal{N}) \leq \liminf\limits_{N\rightarrow \infty}  \E _{N, \omega}  [ u_{\bm {X}_{N}(\omega), \text{loc}} ]\, ,
\end{align}
and $u_{\bm {X}_{N} (\omega), \text{loc}} \to u_{\omega, \text{loc}}^\mathcal{N}$.  The following result provides  important information in practical scenarios. Its proof is a simple adaptation of the previous analysis, upon noticing that Proposition \ref{Proposition:liminf_N} still holds, without assuming that the original sequence consists of global minimisers. 
\begin{Proposition}[\sc limit behaviour of local minimisers as $N\to \infty$]
\label{Thm:min_convergence_prob}
Let $\omega\in \varOmega$ and $V _{\mathcal{N}}$  a fixed DNN space. Consider a sequence of    local minimisers of 
\begin{equation}\label{prob_E_def_theorem}
	\min  _ {v \in V _{\mathcal{N}} } \E _{ {N } , \omega} (v)
\end{equation}
  denoted by   $ u_{\bm {X}_{N} (\omega), \text{loc}}  ,$ where $ \bm{X} _{N} = (X_1, \dots, X_{N}  )  \, .$
Assume that there exists $M(\omega) $ such that 
\begin{equation}\label{w-conv-u_n_bound0_lm}\begin{split} 
   \E _{N, \omega}  \big( u_{\bm {X}_{N} (\omega), \text{loc}} \big) \leq M (\omega)<+\infty\, ,
   \qquad \ \text{for }\ \mathbb{P}-\text{a.e. } \omega\in\varOmega \, .
\end{split}
\end{equation}	
 Then, for $\mathbb{P}$-a.e.\ $\omega\in\Omega$, the sequence $(u_{\bm {X}_{N} (\omega), \text{loc}} )_N$   has weak limits along subsequences with respect to $ (\mu _{N, \bm{X}_{N} (\omega  )}, \mu) $, and for any such weak limit, $u_{\omega, \text{loc}}^\mathcal{N},$  the following estimate holds
\begin{equation}\label {est_final_lm}
  \|u_{\omega, \text{loc}}^\mathcal{N} -f\|_p   \leq  \liminf\limits_{N\rightarrow \infty}  \E _{N, \omega}  \large( u_{\bm {X}_{N}(\omega), \text{loc}} \large)\, , \quad  \text{for }\ \mathbb{P}-\text{a.e. } \omega\in\varOmega
 \, .
\end{equation}
 Furthermore, if the loss corresponding to $u_{\bm {X}_{N}(\omega), \text{loc}} $ has a limit, 
 \begin{equation}\label {est_final_lm_2}
 \lim \limits_{N\rightarrow \infty}  \E _{N, \omega}  \large( u_{\bm {X}_{N}(\omega), \text{loc}} \large)= \beta _{\omega, \text{loc}}^\mathcal{N} \, ,
\end{equation}
then
 \begin{equation}\label {est_final_lm_3}
  \|u_{\omega, \text{loc}}^\mathcal{N} -f\|_p   \leq  \beta _{\omega, \text{loc}}^\mathcal{N} \quad  \text{for }\ \mathbb{P}-\text{a.e. } \omega\in\varOmega
 \, .
\end{equation}
\end{Proposition}
\begin{remark}
One of the main concerns in the implementation of the model machine learning algorithms is that the output of the numerical optimisation step is not a global minimiser. 
The above result, provides conditions, based solely on the loss of these local minima, under which the errors to the target function could be still small. One of these, is the consistent output of small losses for large values of samples. This is yet another demonstration of the robustness of the algorithm.
\end{remark}

\subsection{The $|\mathcal{N}|\to\infty$ limit}\label{calNtoinfty}
In this section we assume the existence of a universal approximation theorem that ensures we may approximate $f$ in $L_p(D,\mu) $ through networks of given architecture and increasing complexity. To avoid confusion,  we select a sequence of spaces  $V _{\mathcal{N}}  $. Thus, for each 
  $\ell \in \mathbb N$ we correspond a DNN space  $ V _{\mathcal{N}}  ,$
  which we call $V_\ell $ with the following property: For each $w\in L^p (D, \mu) \cap S$ 
  there exists a $w_\ell \in V_\ell$ such that,
    \begin{equation}\label{w_ell_7}\begin{split} 
 \|w_{\ell}-w\|_p \leq \  \beta _\ell \, (w), \qquad 
 \text{and } \ \beta _\ell \, (w) \to 0,  \  \ \ell\to \infty\, .	\end{split}
\end{equation}
This assumption is very reasonable in view of the approximations results of neural network spaces, 
see for example \cite{Dahmen_Grohs_DeVore:specialissueDNN:2022,Schwab_DNN_constr_approx:2022, Schwab_DNN_highD_analystic:2023, Mishra:appr:rough:2022, Grohs_Petersen_Review:2023} 
and their references. 
Without loss of generality, we may further assume that 
\begin{equation}
\label{sum_beta}
\sum_\ell \big(\beta _\ell \, (f)\big)^p <+\infty.
\end{equation}
The number of samples we consider for the discrete minimisation problem in $V_\ell$ is denoted by $N(\ell)$ may or may not tend to infinity, as $\ell\to\infty$. 
\begin{Theorem}\label{calNtoinfty_thm}
Consider a sequence of $(V_\ell)_\ell$ of DNN spaces, that satisfies \eqref{w_ell_7} and \eqref{sum_beta}. For each $\omega \in \varOmega$
we consider  the minimiser of the problem 
\begin{equation}\label{prob_E_def_7}
	\min  _ {v \in  V _{\ell} } \E _{ {N(\ell)} , \omega} (v)
\end{equation}
which is denoted by   $ u_{\bm{X}_{N(\ell)} (\omega)   }  ,$ where $ \bm{X} _{N(\ell)} = (X_1, \dots, X_{N(\ell)}  )  \, .$ Then,
\begin{equation}
\label{calNrecovery}
\lim_{\ell\to\infty} \E _{ {N(\ell)} , \omega} ( u_{\bm{X}_{N(\ell)} (\omega)   } ) =0,\qquad \text{for }\ \mathbb{P}-\text{a.e. } \omega\in\varOmega .
\end{equation}
If, in addition, $N(\ell)\to\infty$ and $p>1$, then 
\begin{equation}\label{recovery}
\lim_{\ell\to\infty} u_{\bm{X}_{N(\ell)} (\omega)   } =f,\qquad \text{for }\ \mathbb{P}-\text{a.e. } \omega\in\varOmega,
\end{equation}
in the sense of strong convergence in $L_p$ with respect to $(\mu_{N(\ell),\bm{X}_{N(\ell)}},\mu).$
\end{Theorem}

\begin{proof} Recall that   \eqref{prob_E_min_prop3} implies,
\begin{equation}
\label{prob_E_min_prop_14}
\EE \Big[ \frac 1N  \sum _{i=1}^N   \big  | u_{\bm{X}_{N(\ell)}  }(X_i \, ) - f(X_i   \, ) \big  | ^p \Big]
 \leq  \|w_\ell-f\|_p^p \leq (\beta_\ell(f))^p\, .
\end{equation}
By Markov's inequality we get for every $\epsilon>0$,
\begin{equation}
\mathbb{P}\big[ \E _{ {N(\ell)} , \omega} ( u_{\bm{X}_{N(\ell)} (\omega)   } ) >\epsilon\big] \leq \frac{1}{\epsilon}\mathbb{E}\big[ \E _{ {N(\ell)} , \omega} ( u_{\bm{X}_{N(\ell)} (\omega)   } ) \big] \leq \frac{1}{\epsilon} (\beta_\ell(f))^p.
\end{equation}
In view of \eqref{sum_beta}, the Borel-Cantelli lemma gives that
\begin{equation}
\mathbb{P}\Big[\limsup_{\ell\to\infty} \
 \E _{ {N(\ell)} , \omega} ( u_{\bm{X}_{N(\ell)} (\omega)   } ) =0
\Big] =1,
\end{equation}
which proves the first assertion. If we also have $N(\ell)\to\infty$, by Lemma \ref{GC} we have that $\mu_{N(\ell),\bm{X}_{N(\ell)}(\omega)}\to\mu$, for $\mathbb{P}$-a.e. $\omega$, the result of \eqref{calNrecovery} and triangular inequality together imply condition \eqref{strongconv}, and condition \eqref{w-conv-u_n0} follows by uniform integrability, since $p>1$.
\end{proof}

\begin{remark}[\sc loss for fixed number of samples] 
A clear, but nevertheless very interesting, consequence of
\eqref{prob_E_min_prop_14} and of \eqref{calNrecovery} is that the loss converges to zero as $\ell \to \infty$, even for fixed number of samples. In order to recover $f$ we 
need to take the limit of samples to infinity as well,   \eqref{recovery}.
\end{remark}
\begin{remark}[\sc weak convergence] 
One can show the weak convergence of the entire sequence to $f,$  with respect to $(\mu_{N(\ell),\bm{X}_{N(\ell)}},\mu), $ under the  assumption of \eqref{w_ell_7}, by appropriately modifying the analysis of the Section 
\ref{sec:compactness_convergence}. 
\end{remark}
{\bf Achnowledgements:} ML has been supported by the Hellenic Foundation for Research and Innovation (H.F.R.I.) under the “First Call for H.F.R.I. Research Projects to support Faculty members and Researchers and the procurement of high-cost research equipment grant,” project HFRI-FM17-1034. CM would like to thank V.\ Panaretos, G.\ Savar\' e and J. Xu for 
useful discussions and suggestions. 

\begin{comment}

\textcolor{red}{\bf{This is THE END of the article in the form we discussed in your office. The final section of the previous version follows.}}

\section{Convergence via  liminf - limsup  framework}
In this section we establish the convergence of deep neural network interpolants to the target function $f$ under minimal assumptions. 
A key tool in our approach is De Giorgi's liminf - limsup framework, commonly used in the    $\Gamma$-convergence of functionals, adapted in our present probabilistic setting. The discrete functionals considered are interpreted as before via empirical measures.  
As is typical, the proof consists of the following parts:  we show
the $\liminf$ inequality which provides a lower bound of the discrete energies by the continuum counterpart a.s.\ for $\omega \in \varOmega .$ Then we establish a 
  $\limsup$ inequality which crucially removes randomness from upper bounds and eventually permits to show that the limit of minimisers is, a.s.\ with respect to $\omega ,$ the original function $f.$  
  
  In the sequel,  for reasons related to handling of sequences and subsequences we need to select a sequence of spaces  $ V _{\mathcal{N}}  $ in order to avoid confusion. Thus for each 
  $\ell \in \mathbb N$ we correspond a DNN space  $ V _{\mathcal{N}}  ,$
  which we call $V_\ell $ with the following property: For each $w\in L^p (D, \mu) \cap S$ 
  there exists a $w_\ell \in V_\ell$ such that,
  \cite{Dahmen_Grohs_DeVore:specialissueDNN:2022,Schwab_DNN_constr_approx:2022, Schwab_DNN_highD_analystic:2023, Mishra:appr:rough:2022, Grohs_Petersen_Review:2023} 
  \begin{equation}\label{w_ell}
 \|w_{\ell} - w\|_p \leq \  \beta _{\ell} \, (w), \qquad 
 \text{and } \ 
  \beta _{\ell} \, (w) \to 0,  \  \ \ell \to \infty\, .	
\end{equation} 
To be consistent with the notation of discrete energies
  for each $\omega \in \varOmega$ the minimiser of the problem 
\begin{equation}\label{prob_E_def}
	\min  _ {v \in  V _{\ell} } \E _{N(\ell), \omega} (v)
\end{equation}
is  denoted by   $ u_{\bm{X}_{N(\ell)} (\omega)   }  ,$ and $ \bm{X} _{N(\ell)} = (X_1, \dots, X_{N(\ell)}  )  \, .$
 
\subsection{The $\liminf$ inequality}
Our goal in this section is for a.s.\  $\omega \in   \varOmega $ is to show the existence of a function $u_\omega \in L^p (D, \mu)$ such that 
\begin{align}
 \E (u_\omega) \leq \liminf\limits_{\ell\rightarrow \infty}  \E _{N(\ell), \omega}  [ u_{\bm {X}_{N(\ell)}(\omega)} ]\, ,
\end{align}
and $u_{\bm {X}_{N(\ell)} (\omega)} \to u_\omega$ in an appropriate sense. 
 To achieve this goal, one has to use the weak convergence of measures and relative compactness. However, due to the special structure of the measures
 $$ \big [u_{\bm {X}_{N(\ell)}(\omega)}-f\, \big ]_\sharp\mu _{{N(\ell)}, \bm{X} _{N(\ell)}(\omega)}\, , $$
 this is a subtle task. 

 We introduce first the notion of weak convergence of measures: A sequence of measures $\nu_n $ on $D$ \emph{narrowly (weakly) converges} to $\nu$ if for all bounded and continuous functions, $g : D \to \R$ we have 
 \begin{equation}\label{narrow-conv}\begin{split} 
\lim _{n\to \infty} \int _D \,  g (x) \, \d \nu_n (x) = \int _D \,  g (x) \, \d \nu (x) \, . 	
\end{split}
\end{equation}
Furthermore, if \eqref{narrow-conv} holds,   we say that a sequence of functions $v_n \in L^1 (D, \mu)$ weakly converges to 
$v  \in L^1 (D, \mu)$ with respect to $(\nu_n, \nu), $ if 
 \begin{equation}\label{w-conv-u_n}\begin{split} 
\lim _{n\to \infty} \int _D \, \zeta (x) v_n(x) \, \d \nu_n (x) = \int _D \,  \zeta (x) v(x)  \, \d \nu (x) \, , 
\end{split}
\end{equation}
for all $\zeta \in C_0^\infty (D)\, ,$ see \cite[p.127]{AGS:2005}\,. 
The proof of the following result is an adaptation to our present setting 
of \cite[Theorem 5.4.4]{AGS:2005} and of a classical argument of the weak convergence of empirical measures, \cite{V-58}. 

\begin{Proposition}[\sc the $\liminf$ inequality]The following hold
\label{Proposition:liminf}
\begin{enumerate}
	\item [(i)] There exists a subset $\varOmega _1\subset \varOmega   $ with $\P ( \varOmega \backslash  \varOmega _1)=0$ such that $ \mu _{\ell, \bm{X}_\ell (\omega  )}$ weakly converges to $ \mu$, for all $\omega \in \varOmega _1.$ 
	\item  [(ii)] Let $p>1\, .$ Assume that there exists a   subset $\varOmega _2\subset \varOmega   $ with $\P ( \varOmega \backslash  \varOmega _2)=0$ such that
\begin{equation}\label{w-conv-u_n_bound}\begin{split} 
   \E _{\ell, \omega}  [ u_{\bm {X}_\ell (\omega)} ] \leq M (\omega)\, ,
   \qquad \text{ for all  }\ \omega \in \varOmega _2.
\end{split}
\end{equation}	
Then for all $\omega \in \varOmega _1 \cap \varOmega _2$ there exists a $u_\omega \in L^p (D, \mu)$ such that $u_{\bm {X}_{N(\ell)} (\omega)} $ converges weakly,  {up to a subsequence}, to $u_\omega$ with respect to $ (\mu _{\ell, \bm{X}_{N(\ell)} (\omega  )}, \mu) .$ Furthermore, 
\begin{equation}\label{w-conv-u_n}\begin{split} 
\E (u_\omega) \leq \liminf\limits_{\ell\rightarrow \infty}  \E _{{N(\ell)}, \omega}  [ u_{\bm {X}_{N(\ell)} (\omega)} ]\, .
\end{split}
\end{equation}	
\end{enumerate}
\end{Proposition}

\begin{proof}
Without loss of generality we assume that $N(\ell)=\ell$ in this proof. For completeness we provide the main steps of the proof. For detailed justification  we refer to \cite{AGS:2005}.
 The statement (i) is an straightforward application of Glivenko-Cantelli theorem, \cite{Kallenberg_book}, and follows from the fact that there exists a fixed countable subset $\mathcal {G}$ of the set of bounded continuous functions on $D$ for which it suffices to check the validity of \eqref{narrow-conv}. For each such $g\in \mathcal {G},$ we have 
\begin{equation*} \begin{split}
  \frac 1\ell  \sum _{i}   \Big  | g (X_i(\omega)\, ) - f(X_i (\omega) \, ) \Big  | ^p   \,   \to  \int_D|g(x)-f(x)|^p\, \d \mu(x)\, , \qquad \text{a.s.}, 	
\end{split}
\end{equation*}
by the strong law of large numbers. Since any countable union of sets of measure zero has measure zero, there exists a set $\varOmega _1$
such that (i) holds, \cite{V-58} and \cite{Kallenberg_book}. 
Assumption \eqref{w-conv-u_n} implies that 
\begin{equation}\label{energy_emp_meas}
	\E _{\ell, \omega}  [ u_{\bm {X}_\ell (\omega)} ] = \int _D |u_{\bm {X}_\ell (\omega)} (x)-f(x)|^p\, \d \mu _{\ell, \bm{X}_\ell (\omega)} (x) \leq M (\omega ),  \quad \omega \in \varOmega _2.
\end{equation}
This implies that the sequence of measures $(\big [u_{\bm {X}_\ell (\omega)} -  f \big ]_\sharp \mu _{\ell, \bm{X}_\ell (\omega)} )=
((v_\ell (\omega))_\sharp \mu _{\ell, \bm{X}_\ell (\omega)}) $ is relatively compact. It is not obvious however that the limit points of this sequence of measures will be of the form $ v (\omega)_\sharp \mu \, .$ 
To this end, following, \cite[Theorem 5.4.4]{AGS:2005}, define the plans  $\bm{\gamma} _\ell  = (\bm{i} \times  {v} _\ell )_\sharp \mu _{\ell, \bm{X}_\ell }\ . $  The first marginal of $\bm{\gamma} _\ell $ is 
$\pi ^1 _\sharp \bm{\gamma} _\ell = \mu _{\ell, \bm{X}_\ell} $ and the second, 
$\pi ^2 _\sharp \bm{\gamma} _\ell =  {{v} _\ell}_\sharp  \mu _{\ell, \bm{X}_\ell }.$ Since $\pi ^1 _\sharp \bm{\gamma} _\ell = \mu _{\ell, \bm{X}_\ell} $ weakly converges to $\mu$ and $\pi ^2 _\sharp \bm{\gamma} _\ell $ is relatively compact given that  
$({v_\ell (\omega)}_\sharp \mu _{\ell, \bm{X}_\ell (\omega)}) $   is relatively compact, one concludes that $\bm{\gamma} _\ell $ is relatively compact. Let $\bm{\gamma} $ be such that $\bm{\gamma} _{\ell _k} $ weakly converges to $\bm{\gamma} .$
Then, using that $|y | $ is $ \bm{\gamma} _{\ell}$ uniformly integrable,  
\begin{equation}\label{w-conv-u_n_2}\begin{split} 
\lim _{k\to \infty} \int _D  \, \zeta (x) \,v_{\ell_k} (x) \d \mu _{ {\ell_k}, \bm{X}_{\ell_k}(\omega)} (x) 
&= \lim _{k\to \infty} \int _{\R ^d \times \R^d}   \zeta (x)\,  y_1 \,   \d \bm{\gamma} _{\ell _k} (x, y) \\
&= \int _{\R ^d \times \R^d}   \zeta (x)\,  y_1 \,   \d \bm{\gamma}   (x, y)  \, , \\ 
\end{split}
\end{equation}
where $y_1 $ is the first component of $y.$ Notice however that the first marginal of $ \bm{\gamma} $ is $\mu $ and thus 
\begin{equation}\label{w-conv-u_n_3}\begin{split} 
 \int _{\R ^d \times \R^d}   \zeta (x)\,  y_1 \,   \d \bm{\gamma}   (x, y) 
= \int _{D}   \zeta (x)\,     \overline \gamma_ 1(x)    \d  \mu  (x)  \, , \\ 
\end{split}
\end{equation}
where $\bm{\gamma} = (\gamma_1, \ldots \gamma _d) $ and 
$  \overline {\bm{\gamma}}  (x)  = \int _{\R ^d } y  \d \bm{\gamma}_{x} (y)$ is the barycentric projection with respect to the first marginal of $\bm {\gamma} , $ and $  \overline \gamma_ 1$ is the first component of the vector valued $  \overline {\bm{\gamma}}\, , $  \cite[p.\ 126-8]{AGS:2005}. 
Relationship \eqref{w-conv-u_n_2} implies that $  \overline \gamma_ 1$ depends on $\omega .$ We thus set $ u_\omega (x) =    \overline \gamma_ 1 (x) +f (x)\, .$ 
By the weak convergence of $\bm{\gamma} _{\ell _k} $  to $\bm{\gamma}  $ we finally conclude, 
\begin{equation}\label{w-conv-u_n_4}\begin{split} 
  \ \liminf\limits_{\ell\rightarrow \infty}  \int _{\R ^d \times \R^d}   \, | y_1 | ^p\,   \d \bm{\gamma} _{\ell  } (x, y) &\geq
   \int _{\R ^d \times \R^d}    | y_1 | ^p \,   \d \bm{\gamma}   (x, y)  \, , \\ 
   &\geq \int _{D}       |  \overline \gamma_ 1(x)  |^p  \d  \mu  (x) = \E (u_\omega) \, ,
\end{split}
\end{equation}
and the proof is complete. 
\end{proof} 

 \subsection{The $\limsup$ inequality}
We consider now  the $\limsup$ inequality. 
For each $w\in L^p (D, \mu) \cap S$ 
  we have assumed that there exists a $w_\ell \in V_\ell$ such that 
  \begin{equation}\label{w_ell_2}\begin{split} 
 \|w_{\ell}-w\|_p \leq \  \beta _\ell \, (w), \qquad 
 \text{and } \ \beta _\ell \, (w) \to 0,  \  \ \ell\to \infty\, .	\end{split}
\end{equation}
To be consistent with the notation of discrete energies
  for each $\omega \in \varOmega$ the minimiser of the problem 
\begin{equation}\label{prob_E_def}
	\min  _ {v \in  V _{\ell} } \E _{ {\alpha(\ell)} , \omega} (v)
\end{equation}
is  denoted by   $ u_{\bm{X}_{\alpha(\ell)} (\omega)   }  ,$ and $ \bm{X} _{\alpha(\ell)} = (X_1, \dots, X_{\alpha(\ell)}  )  \, .$

\smallskip
Let is consider a fixed $w.$ We are going to use  \eqref{w_ell_2}  
Since 
\begin{equation}\label{prob_E_min_lsup}\begin{split}
\EE \big[ &\frac 1N  \sum _{i}   \Big  |  w_{\ell} (X_i \, ) - f(X_i   \, ) \Big  | ^p \big]\\
& = \int _\varOmega \,   \frac 1N  \sum _{i}   \Big  |  w_{\ell}(X_i(\omega)\, ) - f(X_i (\omega) \, ) \Big  | ^p  \, \d \PP (\omega) \\
& = \int_D| w_{\ell} (x)-f(x)|^p\, \d \mu(x)\, . 	
\end{split}
\end{equation}
By \eqref{w_ell_2} we have 
$$\lim_ {\ell \to \infty} \int_D| w_{\ell} (x)-f(x)|^p\, \d \mu(x)
= \int_D| w  (x)-f(x)|^p\, \d \mu(x)\, , $$
and thus, 
\begin{equation}\label{w_ell_3}\begin{split} 
\limsup_{\ell\rightarrow \infty}  \EE  \big [ \E _{N(\ell), \omega}  [  w_{\ell} ]   \big ] =
\limsup_{\ell\rightarrow \infty} \int_D| w_{\ell} (x)-f(x)|^p\, \d \mu(x)
= \E (w) \, .
 \end{split}
\end{equation}


\subsection{Convergence of Discrete Minimizers}
\label{sec:compactness_convergence}

\bigskip
We can now combine the above results to conclude that sequences of discrete minimisers converge a.s.\
to the global minimiser of the continuous energy, i.e., they eventually recover $f.$

\begin{Theorem}[\sc convergence of discrete  absolute minimisers]
\label{Thm:min_convergence_prob}
Let $\omega\in \varOmega$ and consider a sequence of    absolute minimisers of 
\begin{equation}\label{prob_E_def_theorem}
	\min  _ {v \in  V _{\ell} } \E _{ {N(\ell)} , \omega} (v)
\end{equation}
  denoted by   $ u_{\bm{X}_{N(\ell)} (\omega)   }  ,$ where $ \bm{X} _{N(\ell)} = (X_1, \dots, X_{N(\ell)}  )  \, .$
Then, a.s.\ with respect to $\omega $
the sequence 
$(u_{\bm {X}_{N(\ell)} (\omega)} )_\ell$ converges weakly   to $f$ with respect to $ (\mu _{\ell, \bm{X}_{N(\ell)} (\omega  )}, \mu) .$ 
\end{Theorem}

\begin{proof} 
By combining the analysis above there exists a set $\widetilde \varOmega  \subset \varOmega ,$  with $\P (\varOmega  \backslash \widetilde \varOmega  )=0 $ such that the following hold for each $\omega \in \widetilde \varOmega :$
We clearly have 
\begin{equation}\label{prob_E_theorem_1}
	\E _{ {N(\ell)} , \omega} (u_{\bm{X}_{N(\ell)} (\omega)   }  ) \leq \E _{ {N(\ell)} , \omega} (v)\, , \quad \text {for all } \ v \in  V _{\ell}  \, . 
\end{equation}
Therefore, selecting $w_\ell $ which are  associated to $w=f,$ we have $\E (w )= \E (f ) =0$ and 
 
\begin{equation}\label{w_ell_3}\begin{split} 
\E _{N, \omega}  &  [  w_{\ell} ] ^{1/p}  = \Big | \E (w ) ^{1/p} -  \E _{N, \omega}  [  w_{\ell} ]   ^{1/p}
   \Big |\\
&= \Big |  \|w  -  f\|_p   - \Big \{ \frac 1N  \sum _{i}   \Big  | w_{\ell}(X_i(\omega)\, ) - f(X_i (\omega) \, ) \Big  | ^p \Big \} ^{1/p} \Big | \\
&\leq   \|w_{\ell}-w\|_p  + \Big | \| w_{\ell}  - f\|_p   - \Big \{ \frac 1N  \sum _{i}   \Big  | w_{\ell}(X_i(\omega)\, ) - f(X_i (\omega) \, ) \Big  | ^p \Big \} ^{1/p} \Big |\, ,	\\
 \end{split}
\end{equation}
where the last term converges to zero a.s. with respect to $\omega$ as $\ell\to \infty $ by the strong law of large numbers, \cite{Kallenberg_book}. Therefore \eqref{w_ell_2} implies that assumption 
 \eqref{w-conv-u_n_bound} holds holds true. 
Then Proposition \ref{Proposition:liminf}  implies
the existence of   $u_\omega \in L^p (D, \mu)$ such that $u_{\bm {X}_{N(\ell)}(\omega)} $ converges weakly {up to a subsequence} to $u_\omega$ with respect to $ (\mu _{\ell, \bm{X}_{N(\ell)} (\omega  )}, \mu) $ and  
\begin{equation}\label{w-prob_E_theorem_2}\begin{split} 
\E (u_\omega)  \leq \liminf\limits_{\ell\rightarrow \infty}  \E _{\ell, \omega}  [ u_{\bm {X}_{N(\ell)} (\omega)} ]\, .
\end{split}
\end{equation}	
Then Fatou's Lemma implies 
\begin{equation}\label{w-prob_E_theorem_3}\begin{split} 
\EE  [\, \E (u_\omega) ]&\leq  \  \EE [\, \liminf\limits_{\ell\rightarrow \infty} \E _{\ell, \omega}  [ u_{\bm {X}_{N(\ell)} (\omega)} ]  ]\leq \liminf\limits_{\ell\rightarrow \infty} \EE  [\, \E _{\ell, \omega}  [ u_{\bm {X}_{N(\ell)} (\omega)} ] ]\\
&  \leq  \limsup_{\ell\rightarrow \infty}  \EE [\,  \E _{\ell, \omega}  [ u_{\bm {X}_{N(\ell)} (\omega)} ] ] \\
& \leq  \limsup_{\ell\rightarrow \infty}  \EE [\,  \E _{\ell, \omega}  [w_\ell ]  ]=\E (w)  =\E (f) 
\, .
\end{split}
\end{equation}	
 Therefore $\EE  [\, \E (u_\omega) ]=0 $  and thus $ u_\omega = f $ a.s.  Since all weakly convergent subsequences of $u_{\bm {X}_{N(\ell)}(\omega)} $ can converge only to $f,$ the entire sequence converges to $f$ for all    $\omega \in \widetilde \varOmega.$
\end{proof}

\medskip


In the case where, the spaces are nested, i.e., $ V_ {\ell '}  \subset V_ \ell $ for $\ell \geq \ell ' ,$ one can prove that the loss converges to zero pointwise. This is of importance, given that for ReLU activation function there holds 
$ \sigma (\sigma (x) \, ) = \sigma (x ) $ and thus it is possible to select the additional coefficients of the larger network equal to zero in order to ensure that $ V_ {\ell '}  \subset V_ \ell $ for $\ell \geq \ell ' .$ 
\begin{Proposition}
	[\sc nested $ V_ {\ell } $]
\label {prob_nested}
If  we assume that $ V_ {\ell '}  \subset V_ \ell $ for $\ell \geq \ell ' ,$ 
then in addition to the conclusions of   of Theorem \ref{Thm:min_convergence_prob} we will have that 
a.s.\ with respect to $\omega $
the sequence 
$(u_{\bm {X}_{N(\ell)} (\omega)} )_\ell$ satisfies
\begin{equation}\label{prob_E_prop_nested}
\lim _{\ell\rightarrow \infty} 	 \E _{ {N(\ell)} , \omega} (u_{\bm{X}_{N(\ell)} (\omega)} ) =0\, . 
\end{equation}
 \end{Proposition}
\begin{proof}
As before, we have, 
\begin{equation}\label{prob_E_prop1_1}
	\E _{ {N(\ell)} , \omega} (u_{\bm{X}_{N(\ell)} (\omega)   }  ) \leq \E _{ {N(\ell)} , \omega} (v)\, , \quad \text {for all } \ v \in  V _{\ell}  \, . 
\end{equation}
Now, selecting $w_\ell $ which are  associated to $w=f,$ and we fix an $w_{\ell '}.$ Using the hypothesis on the spaces $V_\ell$  we have that   $w_{\ell '}\in  V _{\ell}$, for all $\ell \geq \ell '.$ Thus, 
 \begin{equation}\label{prob_E_prop1_2}
	\E _{ {N(\ell)} , \omega} (u_{\bm{X}_{N(\ell)} (\omega)   }  ) \leq \E _{ {N(\ell)} , \omega} (w_{\ell '})\, \quad  \text {for all } \ell \geq \ell '.	\end{equation}
	Since  $w_{\ell '}$  is fixed, the right hand side converges to  $\| w_{\ell '} -f \| _p$  with respect to $\omega$ as $\ell\to \infty $ by the strong law of large numbers. 
	Thus, 
\begin{equation}\label{prob_E_prop_nested_3}
\limsup_{\ell\rightarrow \infty} 	 \E _{ {N(\ell)} , \omega} (u_{\bm{X}_{N(\ell)} (\omega)} )\leq \| w_{\ell '} -f \| _p\, . 
\end{equation}	
Since $\ell' $ was arbitrary, we conclude, in view of \eqref{w_ell_2}, that
 \begin{equation}\label{prob_E_prop_nested_4}
\limsup_{\ell\rightarrow \infty} 	 \E _{ {N(\ell)} , \omega} (u_{\bm{X}_{N(\ell)} (\omega)} ) =0\, . 
\end{equation}
and the proof is complete. 
 \end{proof}

\subsection{The $N\to \infty$ limit}
\label{sec:compactness_convergence}

\bigskip
Now, we fix the DNN architecture and the corresponding space $V _{\mathcal{N}}$ and we take the number of samples to infinity. The following result shows that for a.s.\ with respect to $\omega$ the limit $ u _{\omega} ^  {\mathcal{N}}$ exists and furthermore, its generalisation error   satisfies an optimal approximation error bound.   The proof of this result relies on a modification of the previous analysis. 

\begin{Proposition}[\sc limit behaviour of minimisers as $N\to \infty$]
\label{Thm:min_convergence_prob}
Let $\omega\in \varOmega$ and $V _{\mathcal{N}}$  a fixed DNN space. Consider a sequence of    absolute minimisers of 
\begin{equation}\label{prob_E_def_theorem}
	\min  _ {v \in V _{\mathcal{N}} } \E _{ {N } , \omega} (v)
\end{equation}
  denoted by   $ u_{\bm{X}_{N} (\omega)   }  ,$ where $ \bm{X} _{N} = (X_1, \dots, X_{N}  )  \, .$
Then, a.s.\ with respect to $\omega $ the weak limit of 
the sequence 
$(u_{\bm{X}_{N} (\omega)   } )_N$  with respect to $ (\mu _{\ell, \bm{X}_{N(\ell)} (\omega  )}, \mu) $
exists and is denoted by $u _{\omega} ^  {\mathcal{N} }.$ Furthermore the following estimate holds
\begin{equation}\label {est_final}
  \|u _{\omega} ^  {\mathcal{N} } -f\|_p   \leq  \inf_{\varphi   \in  V _{\mathcal{N}}}  \|\varphi-f\|_p
 \, .
\end{equation}
 
\end{Proposition}

\begin{proof} 
We clearly have 
\begin{equation}\label{prob_E_theorem_2_1}
	\E _{ {N } , \omega} (u_{\bm{X}_{N } (\omega)   }  ) \leq \E _{ {N } , \omega} (v)\, , \quad \text {for all } \ v \in  V _{\mathcal{N}}  \, . 
\end{equation}
Fix  $w \in  V _{\mathcal{N}} , $  and observe 
\begin{equation}\label{w_3}\begin{split} 
\E _{N, \omega}  &  [  w   ] ^{1/p}  = \E (w ) ^{1/p} +
 \Big [ \E (w ) ^{1/p} -  \E _{N, \omega}  [  w ]   ^{1/p}
   \Big ]\\
&=  \|w  -  f\|_p  + \Big [  \|w  -  f\|_p   - \Big \{ \frac 1N  \sum _{i}   \Big  [ w (X_i(\omega)\, ) - f(X_i (\omega) \, ) \Big  | ^p \Big \} ^{1/p} \Big ] \\
 \end{split}
\end{equation}
where the last term converges to zero a.s. with respect to $\omega$ as $\ell\to \infty $ by the strong law of large numbers. Hence, 
\begin{equation}\label{w_4}\begin{split} 
\lim _{N\to \infty} \E _{N, \omega}    [  w   ] ^{1/p}  =   \|w  -  f\|_p  \, .	\\
 \end{split}
\end{equation}
Therefore \eqref{prob_E_theorem_2_1} implies that assumption 
 \eqref{w-conv-u_n_bound} holds holds true. By adopting the proof of 
  Proposition \ref{Proposition:liminf} to our case,  we conclude that 
there exists a    $u_\omega  ^  {\mathcal{N} } \in L^p (D, \mu)$ such that $u_{\bm {X}_{N }(\omega)} $ converges weakly {up to a subsequence} to $u_\omega ^  {\mathcal{N} }$ with respect to $ (\mu _{\ell, \bm{X}_{N(\ell)} (\omega  )}, \mu) $ and  
\begin{equation}\label{w-prob_E_theorem_2}\begin{split} 
\E (u_\omega ^  {\mathcal{N} })  \leq \liminf\limits_{N \rightarrow \infty}  \E _{N, \omega}  [ u_{\bm {X}_{N} (\omega)} ]\, .
\end{split}
\end{equation}	
Then we finally obtain, 
\begin{equation}\label{w-prob_E_theorem_3}\begin{split} 
 \, \E (u_\omega ^  {\mathcal{N} })  &\leq  \  \ \liminf\limits_{N \rightarrow \infty} \E _{N, \omega}  [ u_{\bm {X}_{N} (\omega)} ]      \leq  \limsup_{N\rightarrow \infty}   \E _{N, \omega}  [ u_{\bm {X}_{N} (\omega)} ]   \\
& \leq  \limsup_{N\rightarrow \infty}    \E _{N, \omega}  [w  ]  =\E (w)  \, ,
\end{split}
\end{equation}	
and the proof is complete.
\end{proof}

\begin{remark} The above result shows that the limiting points $u_\omega ^  {\mathcal{N} }$ are minimisers of the problem \eqref{mm_nn:abstract}. 
It is reasonable to expect that these problems attain a global minimum, however, depending on $ V _{\mathcal{N}},$ the minimisers might not be unique. The points $u_\omega ^  {\mathcal{N} }$ are such minimisers. 
In the case where \eqref{mm_nn:abstract} has only one   minimiser $u^\star,$
then a.s. with respect to $\omega,$  $u_\omega ^  {\mathcal{N} }= u^\star.$ Furthermore, if  $ V _{\mathcal{N}}$ are chosen to satisfy  \eqref{w_ell},
all these limiting points converge to $f$ as $\ell \to \infty.$

\end{remark}

\begin{comment}
	
\newpage
\section{Deterministic training : convergence}
\label{sec:G_convergence}

In this section we establish the convergence of deep neural network interpolants to the target function $f.$
A key tool in our approach is a simplified version of $\Gamma$-convergence of the discete functionals considered in the Definition \ref{D-dnnint} 
to the continuum energy $ \E $ with the aim to conclude the corresponding convergence of minimisers. To implement this, it will be essential to \emph{reconstruct} the DNN minimisers and thus to connect appropriately the training step to the limiting functional. 
The proof consists of three parts: we first prove equi-coercivity,  
then, we show
the $\liminf$ inequality which provides a lower bound of the discrete energies by the continuum counterpart. 
We conclude with the $\limsup$ inequality which, as we will see in Section \ref{sec:compactness_convergence} ensures the attainment of the limit. 

\subsection{Reconstructions} A key tool in our analysis is an intermediate function which connects the training points with the space  $V _{\mathcal{N}}.$ This is necessary, given the structure of degrees of freedom within a  neural network which, unlike other typical approximation classes, 
its free parameters are remotely placed and not connected to 
functionals that can control norms of the function.    To this end,     assume that the training points and the quadrature rule has the following property: There exists 
a decomposition $\mathcal{T}_h$  of $\Omega$ into elements $T,$ such that each $z\in T, $ and in addition   
\begin{equation}
\label{quadrature_ass}
  w_z = | T| \qquad\, \text {for }  z\in T\,.
  \end{equation} 
We shall use the notation,   $h_T := \text{diam}(T), $ and $\bar h = \max_{T \in \mathcal{T}_h} h_T.$
 We associate with each decomposition  $\mathcal{T}_h$ the   space  of piecewise constant functions 
\begin{equation*}
{\mathbb{V}}_h :=\{\phi\in L^2(\Omega): \ \ \text{for all } \ T\in\mathcal{T}_h: \phi|_T\in \mathbb{P}_{0}(T)\} ,
\end{equation*}
where $\mathbb{P}_\ell (Q)$ is the space of polynomials   of degree at 
most $\ell$ on $Q.$ 
 As a conclusion, the quadrature rule is exact on ${\mathbb{V}}_h: $   %
 \begin{equation}
\label{quadrature_ass_2}
\sum _{z\in K_h} \, w_z \, \varphi(z) = \int _{\Omega} \, \varphi(x) \, \d x , \qquad \text{for all } \varphi \in {\mathbb{V}}_h\, .
\end{equation} 
 We shall need the interpolation operator $I_ {{\mathbb{V}}_h} \, : C(\overline \Omega) \to {\mathbb{V}}_h\, $ defined by 
\begin{equation}
\label{int_def}
 I_ {{\mathbb{V}}_h}   v \, (z) =  v \, (z) \,  , \qquad \text{for all }  {z\in K_h} \, .
\end{equation} 
With the help of $I_ {{\mathbb{V}}_h} $ one may write the 
loss as 
\begin{equation}
\label{E_h_int}
\mathcal{E}_{Q, h}( g )  =  \int _\Omega  \, \Big  | I_ {{\mathbb{V}}_h} \Big (g(x) -     f(x) \Big ) \Big  | ^p \d x \, . 
\end{equation}


\subsection{Weak convergence}

\begin{prop}[Equi-coercivity]
\label{equi-coercivity}
Let $( u_\ell )_{\ell}$ be a sequence of functions in  $V _{\mathcal{N}} $
 such that for a constant $C>0$ independent of $\ell$ it holds that
$$
\mathcal{E}_{Q, h}( u_\ell ) \leq C.
$$  
Then 
\begin{align}
&\|I_ {{\mathbb{V}}_h}  u_\ell\|_{L^p(\Omega)} \le C. \label{coer:dg_seminorm}
\end{align}
  \end{prop}

\subsection{The $\liminf$ inequality}

\begin{Lemma}[The $\liminf$ inequality.]
\label{Theorem:liminf}
For all $   u \in L^p (\Omega)$ and all sequences 
$( u_\ell )\subset  \bigcup _\mathcal{N} V _{\mathcal{N}} $
such that $$ u_\ell   \rightharpoonup    u$$ in $L^p(\Omega)$  it holds that
 \begin{align}
 \E (u) \leq \liminf\limits_{\ell\rightarrow \infty} \E _{Q, h}[ u_\ell ], 
\end{align}
where the $\ell \to \infty$ limit is understood as a $|\mathcal{N}| \to +\infty $ and simultaneously $h\to 0\, .$
\end{Lemma}
\begin{proof}

 We assume there is a subsequence, still denoted by $ u_h $, such 
 that $\E _{Q,h}[ u_\ell ] \leq C$ uniformly in $h$,
  otherwise $\liminf\limits_{\ell \to \infty }  \E _{Q,h}[ u_\ell]= +\infty$. The proof is based on proving the following steps:
\begin{enumerate}[\labelsep=0.2em 1.]
%
%
 \item Since  
 $   u_\ell    \rightharpoonup    u$ in $L^p(\Omega) $
 \item  $ I_ {{\mathbb{V}}_h} ( u_\ell  -f)   \rightharpoonup    (u-f)$ in $L^p(\Omega) $

 \item The term $\int_\Omega |  u - f  |^p$ is convex which
 implies weak lower semicontinuity \cite{dacorogna2007direct}: 
 $$\liminf\limits_{\ell \to \infty}
\int_\Omega |I_ {{\mathbb{V}}_h} ( u_\ell  -f)  |^p \ge \int_\Omega |u-f|^p.$$


\end{enumerate}
We conclude   that $ \E [   u] \leq \liminf\limits_{\ell \to \infty }  \E _{Q,h}[ u_\ell]$.
\end{proof}

 \subsection{The $\limsup$ inequality}

We consider now  the $\limsup$ inequality. Given $   u \in L^p (\Omega)$,
we would like to prove the existence of  a sequence $( u_\ell )  $, of 
functions in the neural network spaces  
such that $   u_{\ell} \rightarrow    u$ in $L^2(\Omega)$ and 
$  \E [   u] \geq \limsup\limits  \E _{Q,h}[   u_{h}]$ as $h \rightarrow 0$. 
The proof is based on the following argument: (i) One may assume first that $u$ is smooth (ii)    by the approximation properties of the $V_{\mathcal{N}} $ space, 
one may construct   a sequence of   functions $   u_\ell$ such that
 $   u_\ell \rightarrow    u$ in $H^1(\Omega). $ 
 Then, for this sequence,  $\norm{ u_\ell  -    u}_{H^1(\Omega)} \rightarrow 0$ and thus $ \lim_{h\rightarrow 0}   \E _{Q,h}[   u_{\ell}] =  \E [   u] $.
 Upon filling the technical points we can  conclude, 
 

 \begin{Lemma}[The $\limsup$ inequality.]
\label{Theorem:limsup}
The following property holds:
 For all $   u \in L^p (\Omega)$, there exists a sequence
   $( u_\ell )_{\ell}$ with $ u_\ell  \in  \bigcup _\mathcal{N} V _{\mathcal{N}}$,  such that 
$ u_\ell  \rightarrow    u$ in $L^p(\Omega)$ and 
\begin{align}
  \E [   u] \geq \limsup\limits_{\ell\rightarrow \infty}  \E _{Q,h}[ u_\ell ]. 
\end{align}
\end{Lemma} 

\subsection{Compactness and Convergence of Discrete Minimizers}
\label{sec:compactness_convergence}

\bigskip
In this section our main task is to use the results of the previous section to show that under some
boundedness hypotheses on $ u_h $, a sequence of discrete minimizers $( u_h )$
converges in $L^2(\Omega)$ to a global minimizer $   u$ of the continuous functional,
Theorem~\ref{Thm:min_convergence}. 
\begin{Theorem}[Convergence of discrete  absolute minimizers]
\label{Thm:min_convergence}
Let $( u_\ell ) \subset \bigcup _\mathcal{N} V _{\mathcal{N}}$ be a sequence
of  absolute minimizers of $ \E _{Q,h}$, i.e.,
\begin{align}
 \E _{Q,h}[ u_\ell ] = \inf_{w_\ell \in  V _{\mathcal{N}}}  \E _h[w_\ell]. 
\label{eq:discrete_min}
\end{align}
If $ \E  _{Q,h}[ u_\ell ]$ is uniformly bounded then, 
\begin{align}
   &I_ {{\mathbb{V}}_h}  u_{\ell} \rightharpoonup    f,  \qquad\text{ in } L^p(\Omega)\, .
\end{align}
\end{Theorem}

\begin{proof}[Proof of Theorem \ref{Thm:min_convergence}]
The uniform bound for the  energies of the discrete minimisers   implies from the equi-coercivity property, 
Proposition \ref{equi-coercivity}, that
\begin{align}
 \norm{I_ {{\mathbb{V}}_h}  u_{\ell}   }_{L^p(\Omega)}   < C,  
\end{align} 
uniformly.  Therefore, \cite{weak_cov_ref},  there exists $   u\in L^p(\Omega)$ 
such that $ I_ {{\mathbb{V}}_h}  u_\ell     \rightharpoonup    u$ in 
$L^p(\Omega) $ up to a subsequence not relabelled.

Next, we   prove that $   u$ is a global minimizer of $ \E , $ and thus $u=f.$ We use the $\liminf$ and $\limsup$
inequalities, Theorems \ref{Theorem:liminf} and \ref{Theorem:limsup} respectively. 
Let $w \in L^p(\Omega)$, then the $\limsup$ inequality implies that  there exist
$w_\ell \in V _{\mathcal{N}}$ such that 
\begin{align}
 w_h \rightarrow  w \text{ in } L^2(\Omega)  \quad \text{and} \quad 
 \limsup_{h \rightarrow 0}  \E _{Q,h} [w_h] \le  \E [  w ].
\end{align}
Therefore, since $ I_ {{\mathbb{V}}_h}  u_\ell     \rightharpoonup    u$in $L^p(\Omega)$  the $\liminf$ inequality 
and the fact that $ u_\ell $ are absolute minimizers of the discrete problems imply that
\begin{align}
 \E [    u ] \le  \liminf_{h \rightarrow 0}  \E _{Q,h}[ u_h ]  \le  \limsup_{h \rightarrow 0}  \E _{Q,h}[ u_h ] 
\le  \limsup_{h \rightarrow 0}  \E _{Q,h}[ w_h] \le  \E [  w ], 
\end{align}
for all $ w \in L^p(\Omega) $. Therefore $   u$ is an absolute minimizer of $ \E $ and thus $u=f\, .$ Since all weakly convergent subsequences of $\{ I_ {{\mathbb{V}}_h}  u_{\ell} \, \}$ can converge only to $f,$ the entire sequence converges to $f.$ 
\end{proof}



\bibliography{ml_bibliography}

%
%
%
%
%
%
%
%
%
%
%
%
%
%
%
%

\end{document}